\documentclass{article}

\usepackage{PRIMEarxiv}
\usepackage[utf8]{inputenc} % allow utf-8 input
\usepackage[T1]{fontenc}    % use 8-bit T1 fonts
\usepackage{hyperref}       % hyperlinks
\usepackage{url}            % simple URL typesetting
\usepackage{booktabs}       % professional-quality tables
\usepackage{amsfonts}       % blackboard math symbols
\usepackage{nicefrac}       % compact symbols for 1/2, etc.
\usepackage{microtype}      % microtypography
\usepackage{lipsum}
\usepackage{fancyhdr}       % header
\usepackage{graphicx}       % graphics
\graphicspath{{media/}}     % organize your images and other figures under media/ folder
\usepackage{svg}
\usepackage{authblk}
\usepackage[paper=portrait,pagesize]{typearea}
\usepackage{multirow} 
\usepackage{makecell}

\usepackage{geometry}
\usepackage{multirow}
\usepackage{booktabs}
\usepackage{makecell}
\usepackage{pdflscape}

\usepackage{svg}
\usepackage{enumitem}
\usepackage{caption}
\usepackage{bm}

\usepackage{float}
\usepackage{amsmath}

\usepackage{xcolor,pifont}
\newcommand*\colourcheck[1]{%
  \expandafter\newcommand\csname #1check\endcsname{\textcolor{#1}{\ding{52}}}%
}
\colourcheck{blue}
\colourcheck{green}
\colourcheck{red}
\colourcheck{orange}

\usepackage{array}
\usepackage{ragged2e}

\usepackage{titlesec}
\titlespacing{\section}{0pt}{\parskip}{-\parskip}
\titlespacing{\subsection}{0pt}{\parskip}{-\parskip}
\titlespacing{\subsubsection}{0pt}{\parskip}{-\parskip}

\usepackage{tcolorbox}
\usepackage{enumitem}

\usepackage{amsmath}
\usepackage{algorithm}
\usepackage{algpseudocode}
\usepackage{amsfonts}
\usepackage{amsmath}
\usepackage{graphicx}
\usepackage{hyperref}

\usepackage{graphicx}
\DeclareGraphicsExtensions{.png}
\pdfpagewidth=\paperwidth
\pdfpageheight=\paperheight

\usepackage{graphicx}
\usepackage{float}
\usepackage{caption}

\usepackage{graphicx,float,caption}

\usepackage{authblk}
\title{QUAD-LLM-MLTC: Large Language Models Ensemble Learning for Healthcare Text Multi-Label Classification}

\author[ ]{\textbf{Hajar Sakai} and \textbf{Sarah S. Lam}}
\affil[ ]{School of Systems Science and Industrial Engineering}
\affil[ ]{State University of University at Binghamton}
\affil[ ]{Binghamton, NY, USA}
\affil[ ]{hsakai1, sarahlam@binghamton.edu}

\begin{document}
\maketitle

\begin{abstract}
The escalating volume of collected healthcare textual data presents a unique challenge for automated Multi-Label Text Classification (MLTC), which is primarily due to the scarcity of annotated texts for training and their nuanced nature. Traditional machine learning models often fail to fully capture the array of expressed topics. However, Large Language Models (LLMs) have demonstrated remarkable effectiveness across numerous Natural Language Processing (NLP) tasks in various domains, which show impressive computational efficiency and suitability for unsupervised learning through prompt engineering. Consequently, these LLMs promise an effective MLTC of medical narratives. However, when dealing with various labels, different prompts can be relevant depending on the topic. To address these challenges, the proposed approach, QUAD-LLM-MLTC, leverages the strengths of four LLMs: GPT-4o, BERT, PEGASUS, and BART. QUAD-LLM-MLTC operates in a sequential pipeline in which BERT extracts key tokens, PEGASUS augments textual data, GPT-4o classifies, and BART provides topics’ assignment probabilities, which results in four classifications, all in a 0-shot setting. The outputs are then combined using ensemble learning and processed through a meta-classifier to produce the final MLTC result. The approach is evaluated using three samples of annotated texts, which contrast it with traditional and single-model methods. The results show significant improvements across the majority of the topics in the classification’s F1 score and consistency (F1 and Micro-F1 scores of 78.17\% and 80.16\% with standard deviations of 0.025 and 0.011, respectively). This research advances MLTC using LLMs and provides an efficient and scalable solution to rapidly categorize healthcare-related text data without further training. 
\end{abstract}


\section{Introduction}
Textual data belongs to the unstructured data category and, therefore, comes with multiple challenges when intended to extract useful information. Natural Language Processing (NLP) is the artificial intelligence field that combines linguistics and machine learning methods to effectively process and analyze large amounts of data expressed in natural language. NLP tasks are known for being tedious, given that conducting them is costly in terms of time and resources. Multiple techniques were developed over time to establish frameworks that enable their automation. Among these tasks, this paper focuses on Multi-label Text Classification (MLTC). This task aims to assign the relevant labels to a text instance/document instead of one category (Nam et al., 2014). However, in the context of healthcare, MLTC presents several challenges. First, there is the high dimensionality that results from the numerous topics (i.e., labels) that can be mentioned in various clinical notes, patients’ comments, or medical reports. The healthcare text datasets can cover a large array of topics, which increase the methodology's complexity. Additionally, this type of text often contains sophisticated terminology that can be challenging to interpret and accurately classify and would, therefore, require the input of multiple individuals. Second, this type of dataset is characterized by being highly imbalanced, and, therefore, the result is a skewed topic distribution. Consequently, the classifier implemented would suffer from bias and, thus, provide inaccurate assignments regarding potentially critical topics that are not frequently mentioned in the documents’ collection. However, a biased classification would only be problematic if supervised learning is feasible. In multiple cases of carrying out an MLTC in healthcare, for instance, medical abstracts or patients’ feedback, no future event is expected. Therefore, no automatic labeling of the textual data will ever be available if not manually done. A cost-effective solution to this problem is to develop an unsupervised learning approach by leveraging Large Language Models (LLMs). Furthermore, if LLMs are involved, careful consideration should be taken regarding the input textual data. The privacy of healthcare data and the need for Health Insurance Portability and Accountability Act (HIPAA) compliance require the de-identification of patient data to protect sensitive information if included. Text mining techniques can be used to develop a patient’s de-identification framework. A fourth challenge that awaits to be overcome in the case of healthcare textual data MLTC is the quality of data. The different piles of text collected often contain abbreviations and might be informal and inconsistent in grammar, terminology, and style. Traditionally, the text would be tokenized and vectorized. However, the variability found in the text can lead to a very large feature space and, therefore, high dimensionality. Additionally, vectorization is characterized by its limited capacity to embed every contextual nuance in the text representation. Resorting to LLMs can also solve this issue because text is the only valid input data, and vectorization would no longer be a concern. 
\\[0.25cm]
Figure 1 illustrates the research problem of automatically classifying unlabeled healthcare documents into predefined topics using an LLM-based approach. The process begins by inputting textual data into the model, which analyzes the content and assigns relevant topics. The synthetically generated example demonstrates the high-level workflow of this approach, which showcases how unlabeled text is converted into insightful topics. This research focuses on developing an LLM-based approach to achieve these objectives, which contributes to automated healthcare MLTC advancements.

\begin{figure}[H]
    \centering
    \includegraphics[scale=0.7]{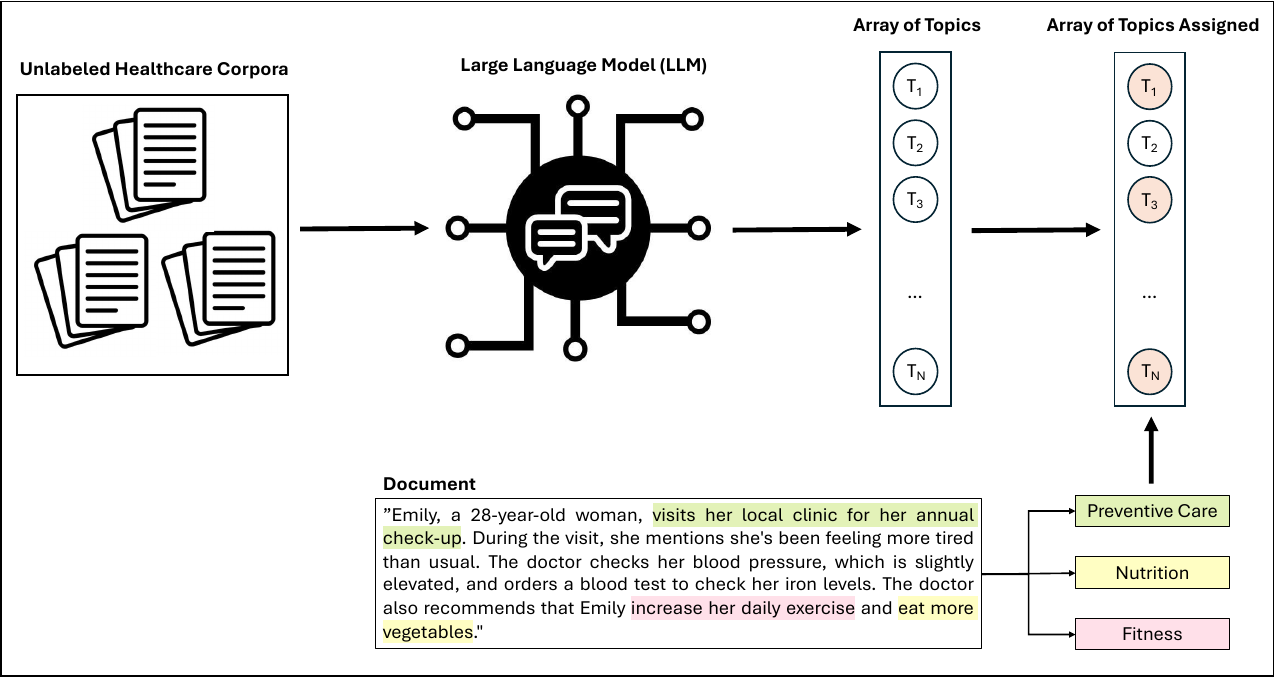}
    \caption*{\textbf{Figure 1.} Research Problem}
    \label{fig:research-prob}
\end{figure}

The introduction of LLMs has revolutionized the way NLP tasks are conducted. Language Models (LMs) have evolved rapidly over the past 30 years. They began as statistical models and then advanced to neural-based LMs. By leveraging the parallel development of computational resources, this progress led to the advent of pre-trained LMs and, ultimately, to the creation of LLMs (Zhao et al., 2024). The evolution of LMs is marked not only by the increasing size of model architectures and the expansion of training data but also by the remarkable enhancement of their task-solving capabilities. LLMs are trained on very large datasets and have been shown to excel in various NLP tasks due to their deep contextual and linguistic understanding (Brown et al., 2020). Among these tasks, LLMs can differentiate and categorize text related to different labels (Loukas et al., 2023; Zhang et al., 2024), which is, if not the same, at least partially the purpose of MLTC. The pre-training of LLMs enables them to capture both the semantic relationships within the text data and the contextual information. This results in context-aware text representations and a better MLTC because both nuances and interdependencies among labels would be considered (Devlin et al., 2018; Yang et al., 2019). Moreover, LLMs are characterized by a robust ability to handle long-range dependencies. This is particularly useful when conducting MLTC, in which different topics can be discussed throughout the document, especially because many healthcare texts are lengthy (Zaheer et al., 2020; Beltagy et al., 2020). Furthermore, LLMs use the knowledge acquired during their pre-training and can generalize without necessarily learning from label-related examples in the case of prompt engineering. As a result, the performance of these tail labels is enhanced without providing many relevant examples (Yin et al., 2019; Chalkidis et al., 2020). This knowledge also enables LLMs to carry out MLTC with limited or no labeled data (Brown et al., 2020), which is crucial in this research. 
\\[0.25cm]
Using LLMs for MLTC can be done in various ways. It can consist of building a task and data-specific LLM from scratch. However, this consumes time and exhausts resources. The use of existing high-performing pre-trained LLMs (e.g., GPT-4o) avoids this issue. Pre-trained LLMs can be used either through fine-tuning or prompt engineering. On the one hand, fine-tuning, as much as it allows quick deployment and task-specific customization, is costly, computationally expensive, can lead to generalization loss, and, most importantly, requires a significant number of good quality examples to learn from, which is quite a challenge in the case of MLTC of healthcare textual data. On the other hand, prompt engineering involves crafting the LLM’s input to maximize output accuracy by providing the required context. It is cost-efficient compared to fine-tuning because savings are made on time and the computational resources used. Additionally, there is no risk of new bias because no retraining is involved. The LLM’s inherent biases are still present but can be controlled through the prompt. Besides, prompt engineering provides both easy implementation and rapid experimentation. 
\\[0.25cm]
The application of LLMs for MLTC in healthcare is currently a very active and promising area of research. This research reveals that prompting an LLM does not necessarily lead to the best performance across all topics. An approach that consists of leveraging four different LLMs, followed by ensemble learning, is introduced to solve this issue. Prompt engineering mainly relies on the amount of context provided in the prompt (Jiang et al., 2020). \( LLM_{1}\) (GPT-4o) is used for the classification and takes the prompt as input. LLM2 (BERT) (Devlin et al., 2018) and LLM3 (PEGASUS) (Zhang et al., 2020) are used to provide more context to the prompt. The former is for key tokens’ extraction, while the latter is for text data augmentation to provide variations of the input text in the prompt. LLM4 (BART) (Lewis et al., 2020) returns topics’ assignment probabilities. 
\\[0.25cm]
This research’s key contributions consist of solutions to the main previously detailed challenges: 

\begin{itemize}
    \item \textbf{Contextual Prompt Engineering:} \( LLM_{2}\) and \( LLM_{3}\) are leveraged to provide more context to the LLM1 prompt by providing the key tokens and text augmentation, respectively. This results in a richer and more informative prompt that leads to a more accurate and relevant classification for some topics.
    \item \textbf{Multimodel Approach:} This approach introduces and evaluates an unsupervised learning approach comprising four distinct LLMs. Each model's unique strengths and capabilities are strategically leveraged to maximize its contributions. Additionally, this approach overcomes the need for both annotated and balanced datasets.
    \item \textbf{Ensemble Learning:} A stacking approach is trained and validated, which takes the binary outputs of three classifications and \( LLM_{4}\) probabilities as input. This enables automated MLTC without requiring the development of an efficient text embedding technique or manually selecting the best approach for each topic.
\end{itemize}

This research uses the Hallmarks of Cancer (HoC) corpus as the data source. This dataset was chosen because experts previously annotated it, and the text included complex medical terminology, which, therefore, can provide evaluation sets. This evaluation was done on three levels to ensure reproducibility and determine the best method, which ensures better overall performance. The three levels consist of three different samples of the dataset (300, 500, and 1,000 texts) and, therefore, different sample sizes and topic distributions. 
\\[0.25cm]
The remainder of the paper is structured as follows: The “Related Literature” section reviews the existing literature about the methods used to multi-label classify healthcare textual data using LLMs. In “Data and Sampling”, the dataset and sampling process are presented. “Proposed Approach” consists of the multi-model ensemble learning approach that enables the MLTC. The “Results and Discussion” section discusses the results of binary, example-based, and level-based classification evaluations. Finally, “Conclusion and Future Directions” provides the research conclusion and summarizes future work.

\section{Related Literature}
Text classification has emerged as an active research area in healthcare, driven by the rapid increase in unstructured data volume and significant advancements in computational capabilities. This task leverages NLP and machine learning techniques to assign labels (i.e., topics) automatically to medical narratives such as clinical notes and diagnosis reports (Wang et al., 2018) or even patients’ comments (Sakai et al., 2023). These applications have shown promise in providing improved decision support and facilitating efficient categorization of large healthcare datasets regardless of the methods used (e.g., unsupervised learning, supervised learning, transfer learning, and rule-based) (Mujtaba et al., 2019).      
\\[0.25cm]
The shift that text classification witnessed from binary to multi-label reflects a significant evolution in NLP. While binary text classification consists of assigning one label per document, MLTC allows assigning multiple labels per document, which better represents real-word textual data (Tsoumakas and Katakis, 2007). This transition enabled capturing more insights from text data. However, it also introduced new challenges, such as label dependency handling and the growing output space (Zhang and Zhou, 2013). The advent of LLMs has changed how healthcare textual data is dealt with, which further advances MLTC and tasks such as medical entity extraction (Ghali et al., 2024). These pre-trained language models have demonstrated promising results by leveraging their deep contextual understanding and transfer learning capabilities across multiple application domains, and healthcare is no exception. Table 1 summarizes relevant research papers on different strategies for MLTC using LLMs in healthcare.
\\[0.25cm]
Following the supervised learning setting, multiple research studies consist of developing a fine-tuning strategy to adapt the general-purpose or domain-specific used LLM to the downstream task (i.e., MLTC) and the considered data. On the one hand, the fine-tuned LLMs are part of the new generation of language models trained on large amounts of data. They are open source and available for research, such as Llama and its variants. Vithanage et al. (2024), for instance, leveraged Parameter Efficient Fine-Tuning (PEFT) to adapt Llama 2-Chat 13B (0-shot and few-shot learning) for free-text nursing progress notes MLTC into four different clinical labels. At the same time, Li et al. (2024) adapted Llama 2 to the clinical domain (LlamaCare) via the Medical Information Mart for Intensive Care III (MIMIC-III) dataset through instruction fine-tuning. LLaMA 13B was also fine-tuned to extract structured condition codes from surgical pathology reports (Path-LLaMA 13B) by Bumgardner et al. (2024). The researchers reported good results in experiments with different dataset sizes. Gema et al. (2023) fine-tuned Llama using Low-Rank Adaptation (LoRA), one of the PEFT strategies, for diagnosis and procedure predictions using MIMIC-IV. On the other hand, the LLMs that are further trained can also be part of the older generation of language models that the literature categorizes as Pre-trained Language Models (PLMs), such as BERT (Zhao et al., 2024). Guo (2024) focused on further training DistilBERT to adapt it to PubMed text data by experimenting with fine-tuning using only the abstracts and both the abstracts and the full texts. Furthermore, Bansal et al. (2023) fine-tuned DeBERTa Large in an attempt to muti-label classify tweets about COVID-19 vaccine concerns. Fine-tuned BERT was also combined with another neural network (i.e., TextRNN) to classify cancer literature at the publication level (Zhang et al., 2023). In some other cases, the PLM selected would be domain-specific or previously fine-tuned and available to use, such as in the case of BiomedVLP-CXR-BERT-General used by Uslu et al. (2024) for radiology reports, which were extracted from MIMIC-CXR, MLTC. Another example is using PubMedBERT and BioMed-RoBERTa-base for MIMIC-III and eICU datasets MLTC, respectively (Yogarajan et al., 2021). On the same note, Guevara et al. (2024) considered adapting Flan-T5 XL and Flan-T5 XXL to extract Social Determinants of Health from Radiotherapy, Immunotherapy, and MIMIC-III Clinic Notes.
\\[0.25cm]
In the literature, LLMs were also part of developing a hybrid approach. Bețianu et al. (2024) proposed DALLMi, a semi-supervised technique for BERT adaptation to new domains with limited labeled data. Among these domains, healthcare was considered through the PubMed dataset. The authors introduced a BERT fine-tuning where a label-balanced sampling was considered, a variational loss per label was computed, and a mix-up regularization was used. Always in the framework of developing a loss function to handle class imbalance, Ray et al. (2023) proposed SH-Focal-BERT. The authors developed a new loss function, Segmented Harmonic Loss, for ClinicalBERT. Furthermore, to deal with the label dependencies by which MLTC is characterized, Chen et al. (2022) developed LITMC-BERT, a transformer-based approach for biomedical literature MLTC. This approach, LITMC-BERT, consists of a shared BERT backbone with a label model to capture label-specific features and a label pair model for label correlations’ modeling. While focusing on the labels, Nguyen and Ji (2021) developed a transformer-based approach, LAME, which incorporates label information into the fine-tuning process of BioBERT by leveraging a label attention layer that helps the model focus on relevant parts of the input text depending on the label. Ge et al. (2023) presented Domain Knowledge Enhanced Classification (DKEC), a domain knowledge enhanced MLTC, to improve performance on few-shot labels. The authors developed group-wise training to handle class imbalance. DKEC also used a label-wise attention mechanism to capture label-specific information. 
\\[0.25cm]
Although the previously discussed approaches are domain-specific and tailored to better understand the data considered for MLTC, they continue to encounter challenges. Because LLMs are based on neural networks, they likely suffer from catastrophic forgetting (Kirkpatrick et al., 2017), in which previous information can be lost when trying to adapt to the new input data. Moreover, fine-tuning is often computationally expensive and resource-intensive (Strubell et al., 2020), which makes it challenging for many organizations, particularly in healthcare, to allocate sufficient budget for this approach. Fine-tuning also requires large amounts of annotated data, usually unavailable for healthcare textual data. To address these issues, some researchers employ LLMs in a 0-shot setting for MLTC. Sushil et al. (2024) used GPT-3.5 and GPT-4 for breast cancer pathology reports MLTC and compared their results to multiple supervised learning models such as Long Short-Term Memory with Attention (LSTM-Att). Their results concluded that GPT-4 outperformed these models. Sakai et al. (2024) investigated conducting multi-label classification of the comments that patients share in the Hospital Consumer Assessment of Healthcare Providers and Systems (HCAHPS) survey. The authors used prompt engineering with GPT-4 Turbo. They contrasted the obtained results from 0-shot and few-shot learning to those from traditional machine learning, BERT, and BART. The 0-shot scenario ended up outperforming the other approaches in terms of the F1 score and its variants. Also interested in conducting an MLTC without the need to annotate training datasets, Zhu et al. (2024) focused on the VaxConcerns dataset to explore LLMs 0-shot classification capabilities. Different prompting strategies were investigated using multiple LLMs. The best performance was achieved using GPT-4 with multipass binary prompting. This strategy involves prompting the LLM while considering one label at a time, and it may lead to a high accuracy but at the expense of increased computational time and resources. On the same note, Sarkar et al. (2023) deduced the good performance of GPT-based MLTC for medical blog articles in a 0-shot setting while prompting through ChatGPT-3.5 and comparing the results of embedding-based multi-label topic inference. Other models’ investigations have also been conducted during the last years; for instance, BART can be mentioned. Raja et al. (2023) explored using LLMs such as BART and fine-tuned BERT for PubMed scientific articles in ophthalmology MLTC, where BART outperformed. BERT’s LLM variants, which were fine-tuned to healthcare data to better handle their nuances and contexts, were used and achieved good results. Examples include Lehman et al. (2023), where BioClinRoBERTa was used for discharge summaries (CLIP) MLTC; Amin et al. (2019), where BioBERT was used for AnimalTestInfo dataset MLTC after translating the text from German to English to automatically assign International Classification of Diseases (ICD)-10 codes to nontechnical summaries; and Yogarajan el al. (2022), in which PubMedBERT was used to MLTC eICU dataset for patient shielding codes prediction. Similar to BERT, other open-source LLMs can be found used in the literature, such as T5, which was employed in a non-autoregressive way (T5Enc) to conduct MLTC for the MIMIC-III and the Biomedical Semantic Indexing and Question Answering (BIOASQ) by Kementchedjhieva and Chalkidis (2023). Furthermore, to leverage multiple models—LLMs and machine learning models—Chaichulee et al. (2022) proposed a majority voting that combines the evaluated models that include mBERT, XLM-RoBERTa, WangchanBERTa, and AllergyRoBERTa.

\begin{landscape}
\begin{table}[h]
\captionsetup{labelformat=empty}
\centering
\caption*{\textbf{Table 1.} MLTC Using LLMs in Healthcare}
\label{tab:classification_models}
\begin{tabular}{c c c c c c c c c c}
 \toprule
\multirowcell{2}[0pt][c]{\centering\makecell{\textbf{Reference}}} & 
\multirowcell{2}[0pt][c]{\centering\makecell{\textbf{Language}}} & 
\multirowcell{2}[0pt][c]{\centering\makecell{\textbf{Data Source}}} & 
\multirowcell{2}[0pt][c]{\centering\makecell{\textbf{Best Approach}}} & 
\multicolumn{6}{c}{\makecell{\textbf{Results of the Best}}} \\
\cmidrule(lr){5-10}
& & &  & \makecell{\textbf{Dataset/} \\ \textbf{Model/Label}} & \makecell{\textbf{Accuracy}} & \makecell{\textbf{F1 score}} & \makecell{\textbf{Precision}} & \makecell{\textbf{Recall}} & \makecell{\textbf{AUC score}} \\

\midrule
\makecell{\textbf{Bețianu et al.} \\ (2024)} & English & PubMed &	DALLMi	& \makecell{PubMed - \\ 50\% available labels} & - &	- & \makecell{Mean Avg: \\ 58.90\%}	& -	& - \\

\midrule
\multirow{4}{*}{\makecell{\textbf{Vithanage et al.} \\ (2024)}} & \multirow{4}{*}{English} & \multirow{4}{*}{\makecell{Free-text \\ Nursing \\ Progress Notes}}  & \multirow{4}{*}{\makecell{0-shot or Few- \\ shot learning \\ with PEFT \\ adaptation of \\ Llama 2-Chat \\ 13B-parameter}}  & \makecell{Agitation in \\ dementia} & 91\%	& 92\% & 91\%	& 90\% &	- \\
\cmidrule(lr){5-10}
 & & & & \makecell{Depression in \\ dementia} & 81\%	& 84\%	& 81\% &	82\% &	- \\
 \cmidrule(lr){5-10}
 & & & & \makecell{Frailty index} & 84\%	& 85\%	& 84\%	& 85\% &	-\\
 \cmidrule(lr){5-10}
 & & & & \makecell{Malnutrition \\ risk factors} & 88\% &	90\%&	88\%&	90\%&	-\\

\midrule
\makecell{\textbf{Sushil et al.} \\ (2024)} & English &\makecell{Breast Cancer \\ Pathology \\ Reports} &	0-shot GPT-4	& -	& -	& \makecell{Avg Macro: \\ 83\%}	& -	&-	& - \\

\midrule
\multirow{2}{*}{\makecell{\textbf{Guo} \\ (2024)}} & \multirow{2}{*}{English} & \multirow{2}{*}{PubMed}  & \multirow{2}{*}{\makecell{Fine-tuned \\ DistilBERT}}  & \makecell{Trained on \\ abstracts only} & 75.15\% &	75.78\% &	76.08\% &	75.84\% &	- \\
\cmidrule(lr){5-10}
 & & & & \makecell{Trained on \\ abstracts + \\ full texts} & 79.85\%	& 74.58\%	& 72.47\%	& 78.45\% & - \\

\midrule
\makecell{\textbf{Zhu et al.} \\ (2024)} & English & VaxConcerns&	\makecell{GPT-4 with \\ multi-pass \\ binary \\ prompting}	& - &	-	& 78.75\% &	- &	-& 	- \\ 

\midrule
\multirow{4}{*}{\makecell{\textbf{Bumgardner et al.} \\ (2024)}} & \multirow{4}{*}{English} & \multirow{4}{*}{\makecell{Surgical \\ Pathology \\ Reports}}  & \multirow{4}{*}{\makecell{Path-LLaMA \\ 13B}}  & \makecell{Tiny Dataset} & 77.80\%	& 77.80\%	& 77.80\%	& 77.80\%	& 82.00\% \\
\cmidrule(lr){5-10}
 & & & & \makecell{Small Dataset} & 72.40\%	& 76.10\%	& 76.50\%	& 76.50\%	& 84.20\% \\
 \cmidrule(lr){5-10}
 & & & & \makecell{Large Dataset} & 74.20\%	& 78.50\%	& 79.30\%	& 78.70\%	& 78.60\%\\
 \cmidrule(lr){5-10}
 & & & & \makecell{Average} & 74.80\%	& 77.50\%	& 77.90\%	& 77.70\%	& 81.60\%\\

\midrule
\makecell{\textbf{Uslu et al.} \\ (2024)} & English & MIMIC-CXR &	\makecell{BiomedVLP- \\ CXR-BERT- \\ General}	& - &	-	& \makecell{Micro:  \\ 80.50\%   \\ Macro: \\ 71.61\%  \\ Weighted: \\80.14\%	} &	\makecell{Micro: \\ 80.27\% \\ Macro: \\ 75.44\%\\ Weighted: \\ 80.22\%	}  &	\makecell{Micro: \\ 80.73\%\\ Macro: \\ 69.98\%\\ Weighted: \\ 80.73\%}  & 	- \\ 

\bottomrule
\end{tabular}
\end{table}
\end{landscape}

\begin{landscape}
\begin{table}[h]
\captionsetup{labelformat=empty}
\centering
\caption*{\textbf{Table 1.} Cont.}
\label{tab:classification_models}
\begin{tabular}{c c c c c c c c c c}
 \toprule
\multirowcell{2}[0pt][c]{\centering\makecell{\textbf{Reference}}} & 
\multirowcell{2}[0pt][c]{\centering\makecell{\textbf{Language}}} & 
\multirowcell{2}[0pt][c]{\centering\makecell{\textbf{Data Source}}} & 
\multirowcell{2}[0pt][c]{\centering\makecell{\textbf{Best Approach}}} & 
\multicolumn{6}{c}{\makecell{\textbf{Results of the Best}}} \\
\cmidrule(lr){5-10}
& & &  & \makecell{\textbf{Dataset/} \\ \textbf{Model/Label}} & \makecell{\textbf{Accuracy}} & \makecell{\textbf{F1 score}} & \makecell{\textbf{Precision}} & \makecell{\textbf{Recall}} & \makecell{\textbf{AUC score}} \\

\midrule
\multirow{6}{*}{\makecell{\textbf{Guevara et al. } \\ (2024)}} & \multirow{6}{*}{English} & \multirow{6}{*}{\makecell{Radiotherapy, \\ Immunotherapy, \\ and  MIMIC-III \\ Clinic Notes}}  & \makecell{Any SDoH: \\ Flan-T5 XXL + \\ Synth. Data}  & \makecell{Radiotherapy \\ Dataset} & -	& Macro: 70\%	& -	& -	& - \\
\cmidrule(lr){4-10}
 & & & \makecell{Any SDoH: \\ Flan-T5 XXL + \\ Synth. Data} & \makecell{Immunotherapy \\ Dataset} & -	& Macro: 71\%	& -	& -	& - \\
 \cmidrule(lr){4-10}
 & & & \makecell{Any SDoH: \\ Flan-T5 XXL} & \makecell{MIMIC-III \\ Dataset} & -	& Macro: 57\%	& -	& -	& -\\
 \cmidrule(lr){4-10}
 & & & \makecell{Adverse SDoH: \\ Flan-T5 XL \\ + Synth. Data} & \makecell{Radiotherapy \\ Dataset} &  -	& Macro: 69\%	& -	& -	& - \\
 & & & \makecell{Adverse SDoH: \\ Flan-T5 XL + \\ Synth. Data} & \makecell{Immunotherapy \\ Dataset} & -	& Macro: 66\%	& -	& -	& - \\
 & & & \makecell{Adverse SDoH: \\ Flan-T5 XL} & \makecell{MIMIC-III  \\ Dataset} & -	& Macro: 53\%	& -	& -	& -\\

\midrule
\multirow{5}{*}{\makecell{\textbf{Li et al. } \\ (2024)}} & \multirow{5}{*}{English} & \multirow{5}{*}{MIMIC-III}  & \multirow{5}{*}{LlamaCare}  & \makecell{Mortality \\ Prediction} & -	& -	& -	& -	& 77.62\% \\
\cmidrule(lr){5-10}
 & & & & \makecell{Length of Stay \\ Prediction} & -	& -	& -	& -	& 68.76\% \\
 \cmidrule(lr){5-10}
 & & & & \makecell{Diagnoses \\ Prediction} & -	& -	& -	& -	& 79.16\%\\
 \cmidrule(lr){5-10}
 & & & & \makecell{Procedures \\ Prediction} & -	& -	& -	& -	& 90.76\% \\
  \cmidrule(lr){5-10}
 & & & & \makecell{Macro Average} & -	& -	& -	& -	& 79.08\%\\

\midrule
\makecell{\textbf{Sakai et al.} \\ (2024)} & English & HCAHPS &	\makecell{0-shot \\
GPT-4 Turbo}	& - &	-	& \makecell{76.12\%  \\ Macro: 72.22\% \\ Micro: 74.43\% \\ Weighted: 73.61\% } &	- &	-& 	- \\

\midrule
\makecell{\textbf{Ray et al. } \\ (2023)} & English & \makecell{MIMIC III and \\ MIMIC IV} &	\makecell{SH-Focal-BERT}	& - &	-	& Micro: 71.28\% &	- &	-& 	- \\

\bottomrule
\end{tabular}
\end{table}
\end{landscape}

\begin{landscape}
\begin{table}[h]
\captionsetup{labelformat=empty}
\centering
\caption*{\textbf{Table 1.} Cont.}
\label{tab:classification_models}
\begin{tabular}{c c c c c c c c c c}
 \toprule
\multirowcell{2}[0pt][c]{\centering\makecell{\textbf{Reference}}} & 
\multirowcell{2}[0pt][c]{\centering\makecell{\textbf{Language}}} & 
\multirowcell{2}[0pt][c]{\centering\makecell{\textbf{Data Source}}} & 
\multirowcell{2}[0pt][c]{\centering\makecell{\textbf{Best Approach}}} & 
\multicolumn{6}{c}{\makecell{\textbf{Results of the Best}}} \\
\cmidrule(lr){5-10}
& & &  & \makecell{\textbf{Dataset/} \\ \textbf{Model/Label}} & \makecell{\textbf{Accuracy}} & \makecell{\textbf{F1 score}} & \makecell{\textbf{Precision}} & \makecell{\textbf{Recall}} & \makecell{\textbf{AUC score}} \\

\midrule
\makecell{\textbf{Raja et al. } \\ (2023)} & English & PubMed &	BART	& \makecell{Clinical \\ Studies Sub- \\ Class} &	-	& Micro: 52\% &	Micro: 49\% &	\makecell{Micro: \\ 61\%} & 68\%\\ 

\midrule
\makecell{\makecell{\textbf{Sarkar et} \\  \textbf{al. }} \\ (2023)} & English & \makecell{Medical Blog \\ Articles} & \makecell{ChatGPT-3.5 \\ (Prompting)} &	-	& - &	60.60\% &	- & - & -\\ 

\midrule
\makecell{\makecell{\textbf{Zhang et} \\ \textbf{al.}} \\ (2023)} & English & \makecell{Dimensions \\ Database} & BERT + TextRNN &	-	& - &	90.34\%	& 93.09\%	& 87.75\%	& -\\ 

\midrule
\multirow{2}{*}{\makecell{\makecell{\textbf{Kementche} \\ \textbf{djhieva and} \\ \textbf{Chalkidis}} \\ (2023)}} & \multirow{2}{*}{English} & \makecell{BIOASQ \\ (PubMed)}  & \multirow{2}{*}{\makecell{T5Enc}}  & \makecell{BIOASQ \\(PubMed)} & -	& \makecell{Micro: 75.1\% \\ Macro: 66\%}	& -	& -	& - \\
\cmidrule(lr){3-3}
\cmidrule(lr){5-10}
 & & \makecell{MIMIC-III} & & \makecell{MIMIC-III} & -	& \makecell{Micro: 60.5\% \\ Macro: 31.1\%} & -	& -	& - \\

\midrule
\makecell{\makecell{\textbf{Lehman et} \\ \textbf{al.}} \\ (2023)} & English & CLIP & BioClinRoBERTa &	-	& - &	\makecell{Micro: 80.5\% \\ Macro: 70.7\%}	& -	& -	& -\\

\midrule
\multirow{2}{*}{\makecell{\textbf{Chen et al.} \\ (2023)}} & \multirow{2}{*}{English} & \makecell{LitCovid \\ BioCreative}  & \multirow{2}{*}{\makecell{LITMC-BERT}}  & \makecell{LitCovid \\ BioCreative} & 80.22\%	& \makecell{F1: 93.84\% \\ Micro: 91.29\% \\ Macro: 87.76\%}	& \makecell{Micro: \\ 85.53\% \\ Macro: \\ 80.48\%}	& -	& - \\
\cmidrule(lr){3-3}
\cmidrule(lr){5-10}
 & & \makecell{HoC} & & \makecell{HoC} & 68.54\%	& \makecell{F1: 91.69\% \\ Micro: 86.48\% \\ Macro: 87.33\%} & \makecell{Micro: \\ 76.97\% \\ Macro: \\ 78.94\%} & -	& - \\

\midrule
\makecell{\makecell{\textbf{Bansal et} \\ \textbf{al.}} \\ (2023)} & English & CAVES & \makecell{Fine-tuned \\ DeBERTa Large} &	-	& - &	Macro:  67\%	& -	& -	& -\\ 

\midrule
\multirow{2}{*}{\makecell{\textbf{Gema et al. } \\ (2023)}} & \multirow{2}{*}{English} & MIMIC-IV  & \multirow{2}{*}{\makecell{LLaMA + Clinical \\ LLaMA-LoRA \\ (Trainable)}}  & \makecell{Diagnoses \\ Prediction} & -	& - & - & -	& 81.97\% \\
\cmidrule(lr){5-10}
 & & & & \makecell{Procedures \\ Prediction} & -	& - & - & -	& 88.69\% \\

\bottomrule
\end{tabular}
\end{table}
\end{landscape}

\begin{landscape}
\begin{table}[h]
\captionsetup{labelformat=empty}
\centering
\caption*{\textbf{Table 1.} Cont.}
\label{tab:classification_models}
\begin{tabular}{c c c c c c c c c c}
 \toprule
\multirowcell{2}[0pt][c]{\centering\makecell{\textbf{Reference}}} & 
\multirowcell{2}[0pt][c]{\centering\makecell{\textbf{Language}}} & 
\multirowcell{2}[0pt][c]{\centering\makecell{\textbf{Data Source}}} & 
\multirowcell{2}[0pt][c]{\centering\makecell{\textbf{Best Approach}}} & 
\multicolumn{6}{c}{\makecell{\textbf{Results of the Best}}} \\
\cmidrule(lr){5-10}
& & &  & \makecell{\textbf{Dataset/} \\ \textbf{Model/Label}} & \makecell{\textbf{Accuracy}} & \makecell{\textbf{F1 score}} & \makecell{\textbf{Precision}} & \makecell{\textbf{Recall}} & \makecell{\textbf{AUC score}} \\

 \midrule
\multirow{8}{*}{\makecell{\textbf{Ge et al.} \\ (2023)}} & \multirow{8}{*}{English} & \multirow{4}{*}{EMS Dataset} & 
\multirow{4}{*}{DKEC-GatorTron} & Head Labels & - & - & 91.80\% & 90.70\% & -\\
\cmidrule(lr){5-10}
& & & & Middle Labels & - &  - & 72.40\% & 71.30\% & - \\
\cmidrule(lr){5-10}
& & & & Tail Labels & - & - & 67.60\% & 67.60\%  & - \\
\cmidrule(lr){5-10}
& & & & Overall &  - & \makecell{Micro: 79.5\%\\Macro: 51.1\%} & 82.20\% & 80.30\% & - \\
\cmidrule(lr){3-10}
& & \multirow{4}{*}{MIMIC-III} & \multirow{4}{*}{DKEC-M-CNN} & Head Labels & - & - & 58.60\% & 61.50\% & - \\
\cmidrule(lr){5-10}
& & & & Middle Labels & - & - & 9.60\% & 49.20\% & -\\
\cmidrule(lr){5-10}
& & & & Tail Labels & - & - & 2.90\% & 19.20\% & -\\
\cmidrule(lr){5-10}
& & & & Overall & - & \makecell{Micro: 55.0\%\\Macro: 4.90\%} & 58.90\% & 54.80\% & -\\

\midrule
\makecell{\makecell{\textbf{Chaichulee } \\ \textbf{et al.}} \\ (2022)} & \makecell{ Thai \& \\ English} & \makecell{Drug Allergy \\ Reports} & \makecell{Majority voting of \\ NB-SVM, \\ ULMFiT, \\ mBERT, XLM- \\ RoBERTa, \\ WangchanBERTa, \\ and \\ AllergyRoBERTa} &	-	& 98.37\% &	98.88\%	& \makecell{P: 99.31\% \\ mAp: 97.07\%}	& 98.72\%	& -\\
 
\midrule
\makecell{\makecell{\textbf{Yogarajan } \\ \textbf{et al.}} \\ (2022)} & English & eICU & PubMedBERT &	-	& - &	\makecell{Micro: 64\% \\ Macro: 39\%}	& -	& -	& -\\ 

\midrule
\makecell{\makecell{\textbf{Nguyen } \\ \textbf{and Ji }} \\ (2021)} & English & HoC & LAME &	-	& - &	83.30\%	& 78.84\%	& 88.63\%	& -\\ 

\midrule
\multirow{2}{*}{\makecell{\textbf{Yogarajan } \\ \textbf{et al.} \\ (2021)}} & \multirow{2}{*}{English} & MIMIC-III & PubMedBERT & \makecell{158 labels - 5 \\ tokens} & -	& \makecell{Micro: 65\% \\ Macro: 41\%} & - & -	& - \\
\cmidrule(lr){3-10}
 & & eICU & \makecell{BioMed- \\ RoBERTa-base}& \makecell{93 labels - \\ 512 tokens} & -	& \makecell{Micro: 60\% \\ Macro: 32\%} & - & -	& - \\

 \midrule
\makecell{\textbf{Amin et al. } \\ (2019)} & \makecell{German \\ (translated to \\ English)} & AnimalTestInfo & BioBERT &	-	& - &	Micro: 73.02\%	& 63.68\%	& 85.56\%	& -\\ 
 
\bottomrule
\end{tabular}
\end{table}
\end{landscape}

\section{Data and Sampling}
Evaluation of MLTC is crucial in the determination of the developed approach's performance and conducting a comparative analysis with existing methods. However, this evaluation cannot be done without labels. For this reason, three subsets were sampled from the publicly available Hallmark of Cancer (HoC) dataset (Baker et al., 2016).

\subsection{Hallmark of Cancer (HoC) Dataset}
During tumor development, cancer cells acquire different biological capabilities called the hallmarks. Considering cancer from the hallmarks’ angle simplifies the complexity of cancer biology. This concept was first introduced in 2000 by Hanahan and Weinberg before being updated in 2011, and it is highly influential in cancer research. There are 10 hallmarks, as shown in Figure 2.

\begin{figure}[H]
    \centering
    \includegraphics[scale=0.5]{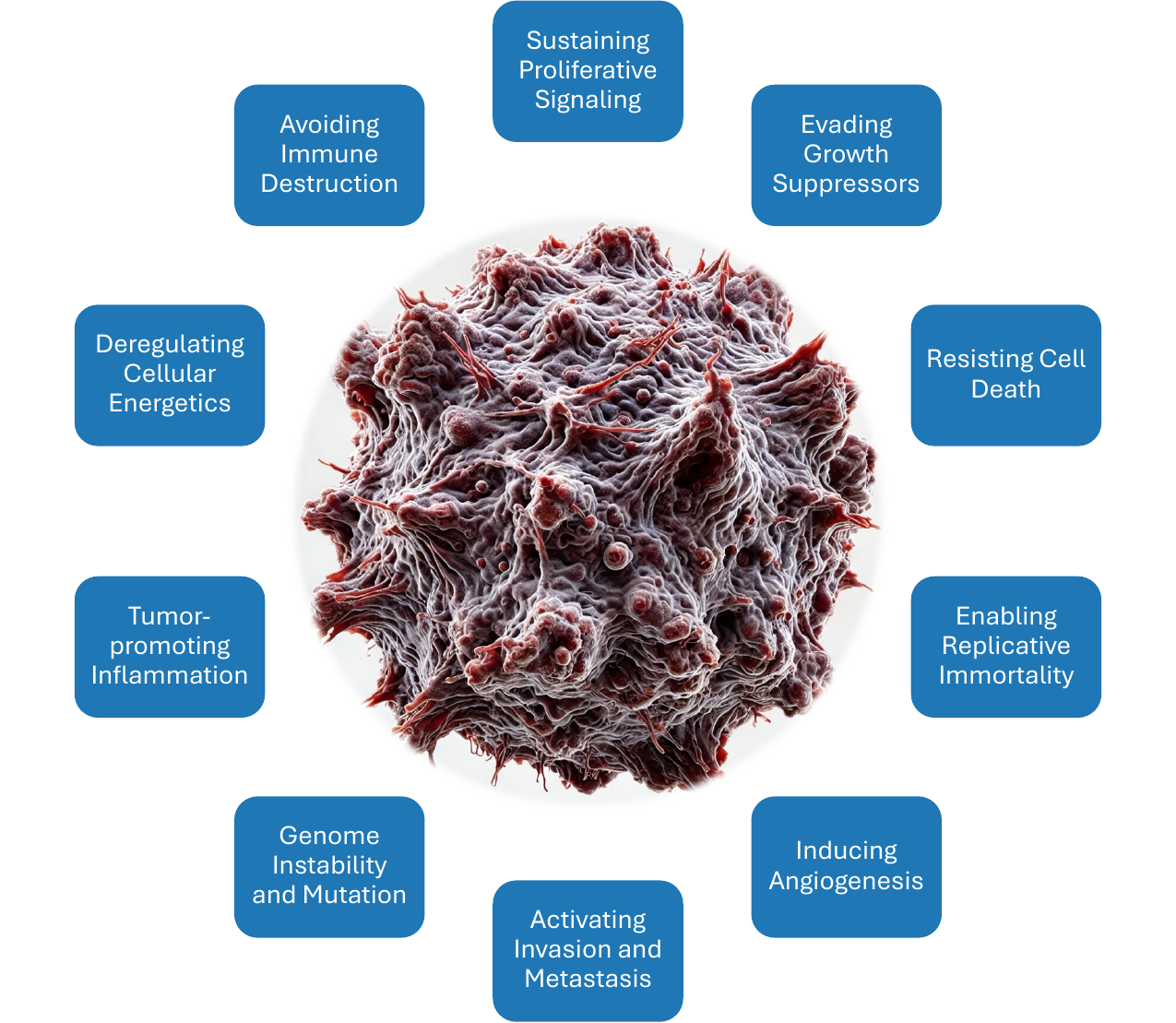}
    \caption*{\textbf{Figure 2.} Hallmarks of Cancer (HoC)}
    \label{fig:mltc-approaches}
\end{figure}

The dataset considered in this research is the Hallmarks of Cancer corpus introduced by (Baker et al., 2016). This corpus consists of 1,499 PubMed abstracts annotated by experts based on the scientific evidence provided for the 10 hallmarks of cancer. The labeling was conducted at a sentence level, in which only sentences that discuss findings or conclusions are considered. Additionally, multiple hallmarks were assigned to the same sentence in case enough evidence was provided. An agreement analysis of this annotation process was conducted on a subset of 155 abstracts (Baker et al., 2016).

\subsubsection{Data Sampling}
To ensure the reliability of the proposed approach, three stratified sets were sampled from the HoC dataset. The three sets were stratified sampled in an attempt to mimic the original dataset topics’ distribution. The sizes of the datasets are as follows: 300, 500, and 1,000, and the topics’ distribution of the largest set is shown in Figure 3. Using different sizes enables the evaluation of how well the proposed approach scales. This is particularly useful when it comes to LLM-based approaches, where performance can depend on the input size, and batch processing is usually employed, given the context size limit. This allows the validity of the prompts’ scalability, because some prompts might behave differently, given whether the input dataset is large or small. Additionally, this iterative evaluation permits the validation of the approach’s generalization capability and sensitivity to the input data volume. Also, it helps to check if the approach’s prompts perform consistently regardless of the amount of data being processed, in addition to how the time and computational resources required scale with the input size.

\begin{figure}[H]
    \centering
    \includegraphics[scale=0.65]{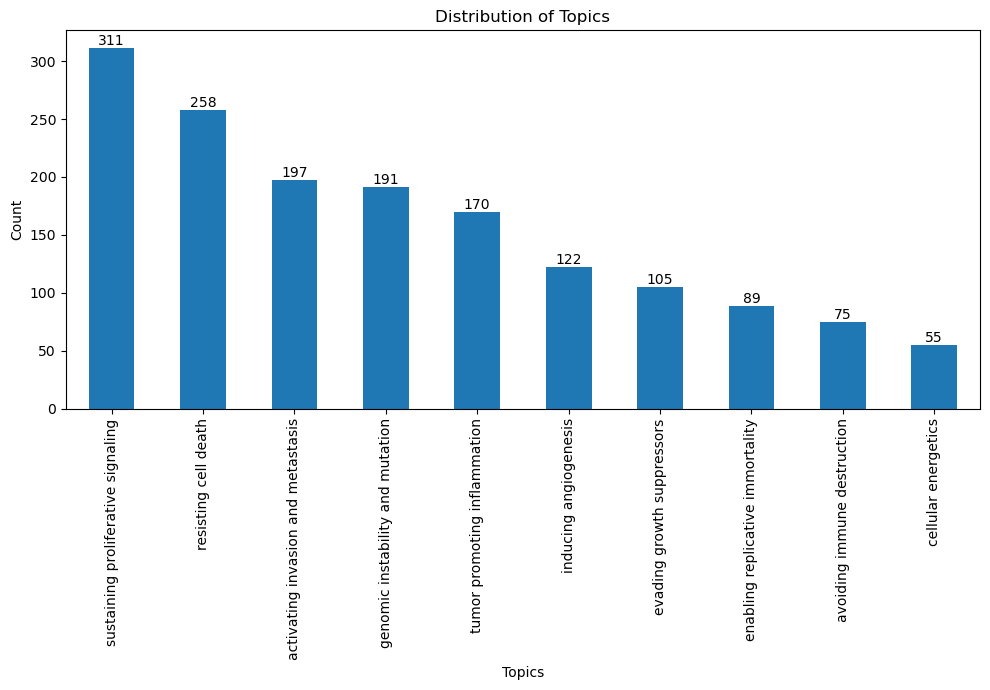}
    \caption*{\textbf{Figure 3.} Topics Distribution}
    \label{fig:mltc-approaches}
\end{figure}

\section{Proposed Approach}
This section presents the proposed approach that leverages four distinct LLMs followed by ensemble learning. Three different prompts are crafted, each with a different context window. The result is three classifications. Furthermore, a fourth LLM is used independently to get the topics’ assignment probabilities, which leads to a fourth classification. These classifications are thereafter stacked and fed as input to a meta-classifier to obtain the final classification output. Each classification was evaluated on three sets of sampled texts with different sizes. As a result, a tailored approach that also ensures consistency is proposed.

\subsection{Problem Definition}
Text classification consists of associating a text (i.e., abstract sentence) with a label (i.e., topic) or more from a set of disjointed labels L. Binary text classification assigns one of two classes, which represents one label, to a text. Multi-class text classification expands this by enabling the assignment of one of multiple classes, all of which represent one label, to a text. However, MLTC allows each text to be assigned multiple labels simultaneously. This hierarchy from binary to multi-class to multi-label text classification shows a significant increase in complexity, and, therefore, a sophisticated approach is required to handle MLTC. For MLTC, given a text t and a label set L, the task consists of determining a subset \( L' \subseteq L \)  that best describes \( t \). The prediction function in MLTC can be expressed as \( f \colon T \to 2^L \), where T is the domain of all texts and \( 2^L \) denotes the power set of labels, which represents all possible label combinations.

\subsection{Large Language Models}
The four LLMs considered for the proposed approach are as follows:

\begin{itemize}[noitemsep, topsep=0pt]
    \item \( LLM_{1}\) is \textbf{GPT-4o}: This LM is part of the \textbf{G}enerative \textbf{P}re-trained \textbf{T}ransformer (GPT) family introduced by OpenAI. It is an enhanced version of GPT-4, the most advanced model from OpenAI at the time of this writing. It is characterized by a large context window that can handle up to 128k tokens in one prompt. Additionally, it is cheaper and faster than GPT-4 and GPT-4 Turbo. Therefore, it is accessible for frequent and large-scale use (OpenAI, 2024a). Furthermore, use of LLM1 involves conducting API requests, which guarantees that the input and output data will not be used for training, which thereby ensures data privacy (OpenAI, 2024b).
    \item \( LLM_{2}\) is \textbf{BERT}: \textbf{B}idirectional \textbf{E}ncoder \textbf{R}epresentations from \textbf{T}ransformers, was introduced by Google Researchers (Devlin et al., 2018), and is particularly characterized by processing the words in a sentence in relation to all the other words instead of ordered one-by-one. As a result, the context understanding of a word is based on all its surroundings. It was pre-trained on a large corpus and can be fine-tuned to various downstream tasks. BERT’s architecture is based on the encoder part of the transformer’s architecture. Thus, one of its key components is the multi-head attention mechanism. The parallel attention heads allow BERT to capture various aspects of semantic and syntactic relationships between words by looking at the input sequence from different representation subspaces. Besides, the special classification token embedding serves as a summary of the entire input sequence. During its pre-training, BERT learned attention weights that depend on the tokens in the training corpus; therefore, leveraging them can be effectively used to extract key tokens. 
    \item \( LLM_{3}\) is \textbf{PEGASUS}: \textbf{P}re-training with \textbf{E}xtracted Gap-sentences for \textbf{A}bstractive \textbf{SU}mmarization \textbf{S}equence-to-sequence was also developed by Google (Zhang et al., 2020) and targets abstractive text summarization. It also uses a transformer-based sequence-to-sequence architecture (encoder + decoder) with a pre-training objective, different from a traditional one, which consists of predicting masked sentences from the text, known as gap sentence generation. Its unique pre-training objective trains the model to reformulate sentences, which makes it well-suited to generate paraphrases. As a result, the LM learns summarization-specific patterns instead of understanding general language. PEGASUS can be adapted for text augmentation through label-preserving paraphrasing. Therefore, variations of the exact text can be provided in the prompt, which enables a better understanding, and is crucial for multi-label classification. 
    \item \( LLM_{4}\) is \textbf{BART}: \textbf{B}idirectional and \textbf{A}uto-\textbf{R}egressive \textbf{T}ransformers, which is also a transformer-based model (Lewis et al., 2020). Like PEGASUS, it involves a bidirectional encoder (similar to BERT) and an autoregressive decoder (similar to GPT), which enables it to understand and generate language effectively. BART’s architecture allows the use of denoising as the pre-training objective, which helps it better handle noisy or incomplete data, a capability relevant to real-world applications. BART can be used for classification by first leveraging its encoder’s outputs to generate contextualized representations of the input text. After that, similar to BERT, the embedding that corresponds to the special classification token can be focused on and fed to a classification layer to predict the probabilities of the different topics. These probabilities help determine the most likely to-be-mentioned topic and assess the model's confidence in its predictions. 
    
\end{itemize}

\subsection{Ensemble Learning Base Classifiers}
The proposed approach consists of an ensemble learning that involves four LLMs and a machine learning classifier. The LLMs (i.e., base classifiers) are leveraged to conduct four classifications in which outputs are fed to the meta-classifier. Below, the crafted prompts and methods used are discussed and summarized. Figure 4 is a detailed representation of the proposed approach. The three crafted prompts are dynamic and change depending on the text under consideration for classification.

\begin{figure}[H]
    \centering
    \includegraphics[scale=0.75]{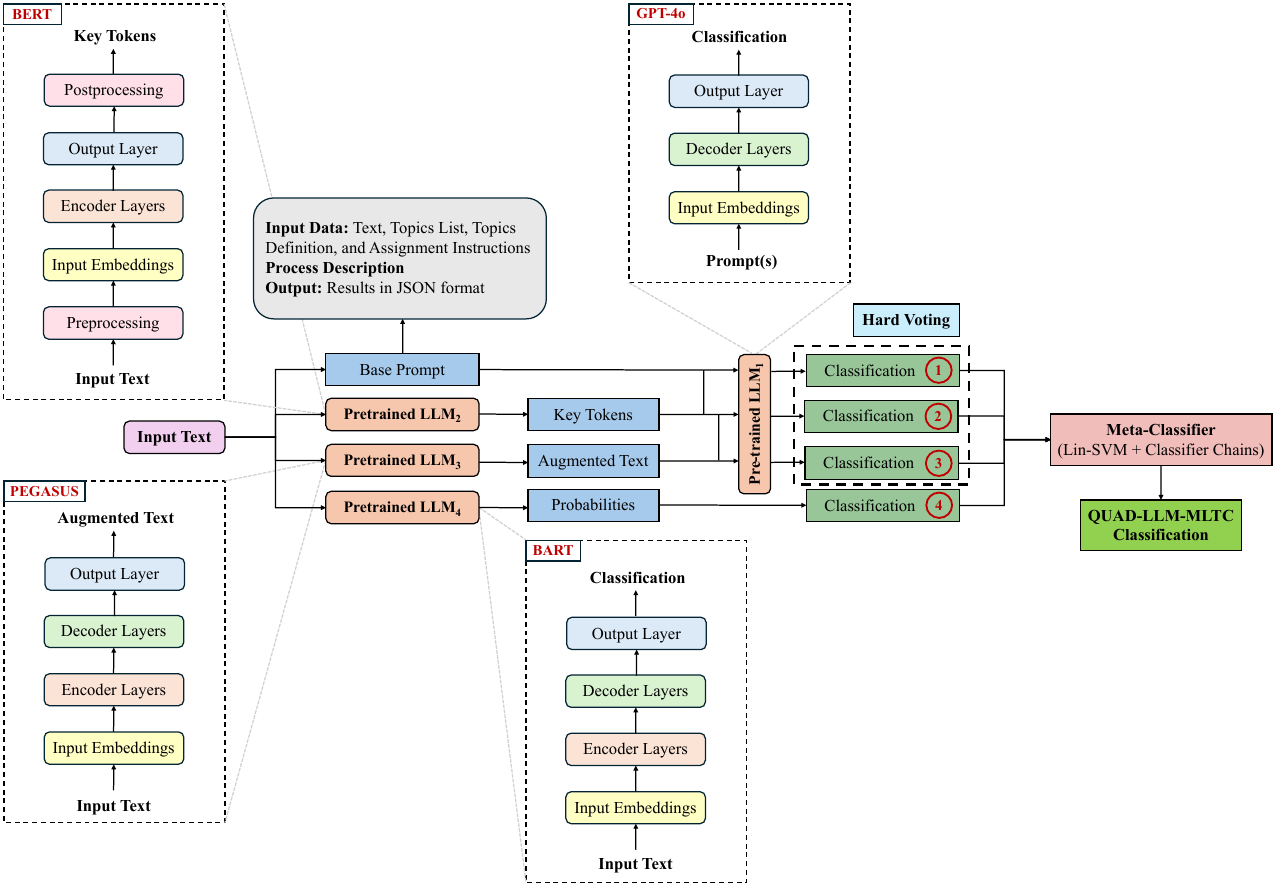}
    \caption*{\textbf{Figure 4.} Proposed QUAD-LLM-MLTC Approach}
    \label{fig:mltc-approaches}
\end{figure}

\subsubsection{Base Prompt}
The first crafted prompt is the \textit{Base Prompt} (Figure 5); it serves as input for \( LLM_{1}\) and includes the abstract’s sentence, the list of the 10 predefined topics, a dictionary with the topics’ definitions, the instructions, the assignment process description, as well as the desired output format. The outputs are \textit{Classification 1}.

\begin{figure}[H]
\centering
\captionsetup{labelformat=empty}
\begin{tcolorbox}[colframe=black, colback=white, coltitle=black, boxrule=0.3mm, sharp corners]
input = f"""{text}""" 
\newline
prompt = f"""You are a healthcare expert, and you want to assign the relevant hallmarks contained in {topics} to each of the texts included in the list delimited by triple backticks. The definition of each hallmark can be found in the dictionary {topics\_details}. Please conduct these assignments without interpretation or extrapolation. Follow the instructions in {instructions}. If no topic is assigned to a text, extract the corresponding one. Provide the results in JSON format with the following keys: Text, Topics. {input}"""
\end{tcolorbox}
\caption{\textbf{Figure 5.} Base Prompt}
\end{figure}

\subsubsection{Base and Key Tokens Prompt}
The second prompt (Figure 6) consists of the Base Prompt and the text’s Key Tokens extracted using BERT. Algorithm 1 shows the details of the extraction approach. The outputs are \textit{Classification 2}.

\newpage

\begin{figure}[H]
\centering
\captionsetup{labelformat=empty}
\begin{tcolorbox}[colframe=black, colback=white, coltitle=black, boxrule=0.3mm, sharp corners]
input = f"""{text}""" 
\newline
prompt = f"""You are a healthcare expert, and you want to assign the relevant hallmarks contained in {topics} to each of the texts included in the list delimited by triple backticks. The definition of each hallmark can be found in the dictionary {topics\_details}. Please conduct these assignments without interpretation or extrapolation. Follow the instructions in {instructions}. The keywords in this text are {key\_tokens}. If no topic is assigned to a text, extract the corresponding one. Provide the results in JSON format with the following keys: Text, Topics. {input}"""
\end{tcolorbox}
\caption{\textbf{Figure 6.} Base Prompt + Key Tokens}
\end{figure}

\begin{algorithm}
\caption{\textbf{: Pseudocode of Key Tokens Extraction Using BERT}}
\textbf{Input:} Text, Top\_k – the number of key tokens to return \\
\textbf{Output:} List of Top\_k key tokens from text based on attention weights
\begin{algorithmic}[1]
\State Initialize BERT tokenizer and model with pre-trained 'bert-base-uncased' parameters, load stopwords
\State Define get\_key\_tokens(Text, Top\_k):
\begin{enumerate}
    \item[a.] Tokenize Text using BERT tokenizer, truncate texts longer than 512 tokens
    \item[b.] Pass the tokenized Text through BERT model and get attention weights
    \item[c.] Average attention weights across all heads using the last layer’s outputs
    \item[d.] Select attention from the classification token [CLS] to represent the Text
    \item[e.] Tokenize the text into lowercase words to match the BERT tokenization
    \item[f.] Sort indices of these tokens in descending order based on their attention weights (token\_indices)
    \item[g.] Initialize set to track seen tokens, ensure uniqueness and relevance of the final key tokens list
    \item[h.] for idx in token\_indices:
    \begin{enumerate}
        \item[i.] Decode each token back to its textual representation and normalize it
        \item[ii.] Check exclusion criteria: not CLS/SEP token, not stopword/subword/number, must be in original Text
        \item[iii.] Add qualified tokens to the list of relevant tokens until top\_k tokens are identified
    \end{enumerate}
    \item[i.] Return list of Top\_k key tokens
\end{enumerate}
\State Remove punctuation from Text, calculate the text's token\_count
\State Based on token\_count:
\begin{enumerate}
    \item[a.] If $\textnormal{token\_count} \leq 50$, get\_key\_tokens(Text, Top\_k = 3)

    \item[b.] If $50 < \textnormal{token\_count} \leq 100$, get\_key\_tokens(Text, Top\_k = 5)
    \item[c.] If $\textnormal{token\_count} > 100$,  get\_key\_tokens(Text, Top\_k = 10\% $\times$ token\_count)
\end{enumerate}
\State Return the list of Top\_k key tokens from Text
\end{algorithmic}
\end{algorithm}

\subsubsection{Base, Key Tokens, and Augmented Text Prompt}

The third prompt (Figure 7) consists of the Base Prompt and the Key Tokens, in addition to two variations of the text generated using PEGASUS. Algorithm 2 shows the details of the text augmentation approach. The outputs are \textit{Classification 3}.

\begin{figure}[H]
\centering
\captionsetup{labelformat=empty}
\begin{tcolorbox}[colframe=black, colback=white, coltitle=black, boxrule=0.3mm, sharp corners]
input = f"""{text}""" 
\newline
prompt = f"""You are a healthcare expert, and you want to assign the relevant hallmarks contained in {topics} to each of the texts included in the list delimited by triple backticks. The definition of each hallmark can be found in the dictionary {topics\_details}. Please conduct these assignments without interpretation or extrapolation. Follow the instructions in {instructions}. The keywords in this text are {key\_tokens}. Additionally, "{variation1}" and "{variation2}" are two variations of the input text and are solely provided to help you better understand it. Do not assign topics to these variations. If no topic is assigned to a text, extract the corresponding one. Provide the results in JSON format with the following keys: Text, Topics. {input}"""
\end{tcolorbox}
\caption{\textbf{Figure 7.} Base Prompt + Key Tokens + Augmented Text}
\end{figure}

\begin{algorithm}
\caption{\textbf{: Pseudocode of Text Data Augmentation Using PEGASUS}}
\textbf{Input:} Text, Num\_return\_variations – number of variations to generate (default = 2), Num\_beams – number of beams in the beam search (default = 5) \\
\textbf{Output:} List of Text's variations
\begin{algorithmic}[1]
\State Import the Pegasus model and tokenizer specifically tuned to paraphrase tasks
\State Initialize Pegasus model and tokenizer with pre-trained parameters from 'tuner007/pegasus\_paraphrase'
\State Define get\_variations(model, tokenizer, Text, Num\_return\_variations, Num\_beams):
\begin{enumerate}
    \item[a.] Tokenize Text using Pegasus tokenizer, applying truncation, and padding for uniform tensor shape
    \item[b.] Generate Text's variations using the model:
    \begin{enumerate}
        \item[i.] Set the number of beams for exploration during generation
        \item[ii.] Specify the number of variations to return
    \end{enumerate}
    \item[c.] Decode the generated token IDs back to text while skipping any special tokens used by the model
\end{enumerate}
\State Return list of Text's variations
\end{algorithmic}
\end{algorithm}

\subsubsection{BART Probabilities}
When tested and evaluated on three different text sets, each of the three prompts produces a classification that can be considered the most robust for one or more topics. However, consistency was not achieved across all topics and all evaluation sets. To tackle this issue, BART is used as LLM4 to provide Classification 4, and an ensemble learning technique is used to enhance the overall performance and robustness of the proposed approach by combining the four classification outputs. However, before that, all the LLMs generated results are postprocessed, and the result is 40 variables, 30 of which are binary and the remaining 10 continuous with values between 0 and 1.

\begin{figure}[H]
    \centering
    \includegraphics[scale=0.75]{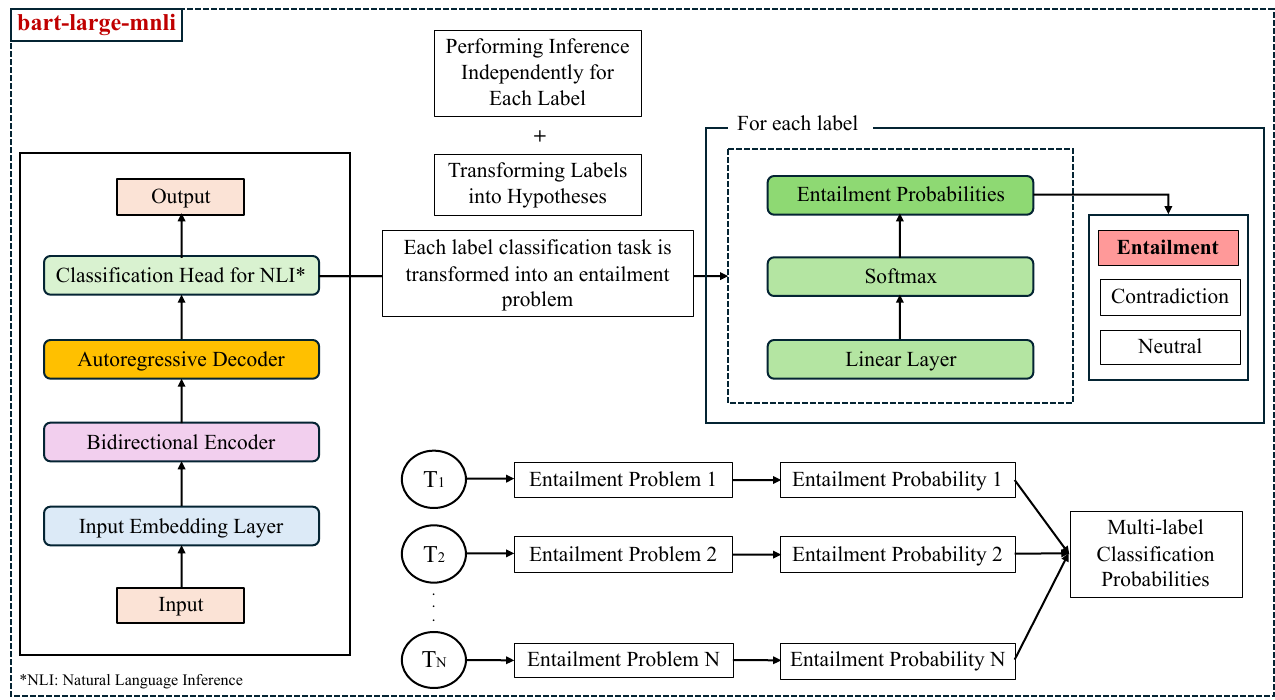}
    \caption*{\textbf{Figure 8.} BART Architecture to Generate Probabilities}
    \label{fig:mltc-approaches}
\end{figure}

\subsection{Ensemble Learning Meta Classifier}
Ensemble learning combines predictions or classifications from multiple models or approaches to achieve better results than any single model alone. This approach reduces overfitting and improves generalization. This technique encompasses several methods. One of the most popular ones is the Bagging (Bootstrap Aggregating) method, where multiple instances of the same algorithm are trained on different subsets, and afterward, their predictions are aggregated. This same method is used for decision trees, which results in Random Forests (Breiman, 2001). Furthermore, Boosting is another popular ensemble method in which weak algorithms (i.e., learners) are trained sequentially, and each new model focuses on the errors generated by the previous ones. Gradient Boosting Machines (Friedman, 2001) and AdaBoost (Freund \& Schapire, 1997) are two popular ensemble boosting algorithms that can be cited. Additionally, there is the Voting technique, where predictions from multiple models are combined using simple mechanisms such as the majority voting in the case of classification (Dietterich, 2000). Moreover, Stacking (Wolpert, 1992) is another ensemble learning method where the combination is done using a meta-classifier (i.e., meta-learner). In this research, the base learners’ outputs considered are those of the previously detailed four text classifications. These outputs are the input features of the meta-learner. 
\\[0.25cm]
This research conducts extensive experiments (i.e., 5x3x3 = 45) to select the best classifier (Lin-SVM vs Logistic Regression vs Random Forest vs Xgboost vs Multi-Layer Perceptron) using the best MLTC transformation technique (Label Powerset vs Classifier Chains vs Binary Relevance) that is consistent across the tree validation subsets. The experiments are carried out using an iterative stratified five-fold cross-validation. All this led to the choice, as mentioned in Figure 4, of a Linear Support Vector Machine (Lin-SVM) using Classifier Chains.

\subsection{Performance Metrics}
The performance of the proposed approach and existing methodologies was assessed by gathering True Positive (TP), True Negative (TN), False Positive (FP), and False Negative (FN) outcomes using the resulting confusion matrix. Because this research problem is an MLTC, different evaluation strategies can assess the approach’s performance: example-based, label-based, and binary evaluations. The \textit{example-based} metrics evaluate the accuracy of the classifications for each text as a whole, given the subset of topics assigned against the actual topics mentioned (Sorower, 2010). For this case, the F1 score is calculated as follows:
\begin{equation}
F1 = 2 \times \frac{ \text{Precision} \times \text{Recall}}{\text{Precision} + \text{Recall}}
\end{equation}
where
\begin{equation}
\text{Precision} = \frac{TP}{TP + FP}
\end{equation}
and 
\begin{equation}
\text{Recall} = \frac{TP}{TP + FN}
\end{equation}
The \textit{label-based} evaluation considers each topic as a separate binary classification problem, calculates the metrics, and then averages them across all topics (Sorower, 2010). In this case, Micro-F1, Macro-F1, and Weighted-F1 scores are calculated as follows:
\begin{equation}
\text{Micro-F1} = 2 \times \frac{ \text{Micro-Precision} \times \text{Micro-Recall}}{\text{Micro-Precision} + \text{Micro-Recall}}
\end{equation}
where
\begin{equation}
\text{Micro-Precision} = \frac{\sum_{j=1}^{|L|} TP_j}{\sum_{j=1}^{|L|} TP_j + \sum_{j=1}^{|L|} FP_j}
\end{equation}
and 
\begin{equation}
\text{Micro-Recall} = \frac{\sum_{j=1}^{|L|} TP_j}{\sum_{j=1}^{|L|} TP_j + \sum_{j=1}^{|L|} FN_j}
\end{equation}
here, \( \text{TP}_j \), \( \text{FP}_j \), and \( \text{FN}_j \) are true positives, false positives, and false negatives for the \( j^{th} \) topic.

\begin{equation}
\text{Macro-F1} = \frac{1}{T} \sum_{j=1}^{|L|} F1_j
\end{equation}
where \( \text{F1}_j \) is the F1 score of the \( j^{th} \) topic.

\begin{equation}
\text{Weighted-F1} = \sum_{j=1}^{|L|} w_j \times F1_j
\end{equation}
where
\begin{equation}
w_j = \frac{\text{Number of instances of topic } j}{\text{Total number of comments}}
\end{equation}

\textit{Binary} evaluation simplifies the evaluation task to multiple independent binary classifications (first-order strategy) (Zhang \& Zhou, 2013), where each topic is assessed individually. The F1 and Area Under the Curve (AUC) scores are calculated in this case. 
\\[0.25cm]
The combination of these evaluation strategies provides a comprehensive summary of the proposed approach and the performance of existing methodologies by addressing the individual topic and overall text levels of classification accuracy.

\section{Results and Discussion}
In this section, QUAD-LLM-MLTC performance is discussed by comparing it against multiple existing approaches that can potentially be used for MLTC. Moreover, an ablation study is carried out to demonstrate the importance of each component of the proposed QUAD-LLM-MLTC approach while comparing the proposed stacking technique with another ensemble learning technique (majority voting).

\subsection{Performance Metrics Summary}
Performance contrasting is based on the evaluation conducted. This latter is done on three levels: a binary classification evaluation, an example-based evaluation, and finally, a label-based evaluation. The evaluation is the primary reason behind sampling sets from an annotated dataset (i.e., HoC). The labeled samples allow for both the training and validation of supervised methods as well as the evaluation of unsupervised approaches.

\subsubsection{Binary Classification Evaluation}
The binary classification evaluation is based on computing the F1 and AUC scores for each of the MLTC topics. The task is treated as if it were transformed into 10 independent binary classifications, in which each classification target is a topic. It is worth noting that the reported performance for Tables 2 and 3 concerns the largest evaluation set of 1,000 texts.

\begin{table}[H]
\captionsetup{labelformat=empty}
\centering
\renewcommand{\arraystretch}{1.2}
\caption{\textbf{Table 2.} Traditional Machine Learning and Pretrained Language Models Results}
\begin{tabular}{c c c c c c c}
\hline
\textbf{Supervised Learning}& \multicolumn{2}{c}{\makecell{\textbf{TF-IDF + Lin-SVM} \\ \textbf{(Label Powerset)}}} & \multicolumn{2}{c}{\textbf{BERT}} & \multicolumn{2}{c}{\textbf{BART}} \\
\hline
 \textbf{Topics/Metrics} & \textbf{F1} & \textbf{AUC} & \textbf{F1} & \textbf{AUC} & \textbf{F1} & \textbf{AUC} \\
\hline
\multicolumn{1}{l}{\textbf{Sustaining proliferative signaling}} & 78.98\% & 84.12\% & 26.40\% & 47.88\% & 27.98\% & 53.17\% \\
\hline
\multicolumn{1}{l}{\textbf{Resisting cell death }} & 89.71\% & 91.57\% & 21.73\% & 48.00\% & 32.48\% & 46.01\% \\
\hline
\multicolumn{1}{l}{\textbf{Activating invasion and metastasis}} & 87.36\% & 89.86\% & 15.29\% & 49.31\% & 58.33\% & 73.04\% \\
\hline
\multicolumn{1}{l}{\textbf{Genomic instability and mutation }} & 84.30\% & 88.88\% & 18.64\% & 50.60\% & 49.82\% & 67.09\% \\
\hline
\multicolumn{1}{l}{\textbf{Tumor promoting inflammation }} & 78.60\% & 82.76\% & 13.19\% & 49.62\% & 30.30\% & 58.51\% \\
\hline
\multicolumn{1}{l}{\textbf{Inducing angiogenesis}} & 82.46\% & 85.54\% & 14.68\% & 52.00\% & 67.66\% & 77.24\% \\
\hline
\multicolumn{1}{l}{\textbf{Evading growth suppressors}} & 61.99\% & 74.51\% & 11.32\% & 51.77\% & 18.48\% & 53.04\% \\
\hline
\multicolumn{1}{l}{\textbf{Enabling replicative immortality}} & 77.03\% & 81.91\% & 7.09\% & 50.23\% & 22.43\% & 56.41\% \\
\hline
\multicolumn{1}{l}{\textbf{Avoiding immune destruction}} & 63.16\% & 73.84\% & 4.92\% & 49.62\% & 16.93\% & 57.17\% \\
\hline
\multicolumn{1}{l}{\textbf{Cellular energetics}} & 72.09\% & 78.18\% & 0.00\% & 48.31\% & 12.36\% & 54.60\% \\
\hline
\end{tabular}
\end{table}

First, three models—TF-IDF with Lin-SVM (Classifier Chains), BERT, and BART—are evaluated to establish a performance baseline before introducing the proposed approach. The traditional model combines Term Frequency-Inverse Document Frequency (TF-IDF), Classifier Chains transformation approach, and Lin-SVM. After standard text preprocessing, TF-IDF was used to transform the texts into numerical feature vectors, and the Classifier Chains to transform MLTC into a series of binary classification problems, which captures label (i.e., topic) dependencies by using previous predictions as features for subsequent classifiers. Lin-SVM, which is a variation of SVM with a linear kernel, was then fitted to train on the TF-IDF vectors and transformed feature space, which aims to optimize class separation. When it comes to traditional methods, other experiments were also conducted. First, the classification method was experimented with, into which Multinomial Naïve Bayes (MNB) was also fitted. Second, multiple problem transformation methods that included Label Powerset and Binary Relevance in the context of MLTC were tried out as well. The results were far from satisfactory and, therefore, not reported. The performance of TF-IDF + Lin-SVM (Classifier Chains) reflects the outputs of a five-fold iterative stratified cross-validation, which is suitable for MLTC. The results demonstrate a relatively good overall performance, especially regarding topics with sufficient training examples, as shown in the topics’ distribution (Figure 3). Furthermore, two Pre-trained Language Models (PLMs) were tested out as is. The first one is BERT, and more specifically, ‘bert-base-uncased.’ BERT underperformed and took around four hours to obtain the iterative stratified cross-validation outputs. This might highlight PLMs' challenges when aligning with specific domains’ textual data, such as healthcare abstracts’ sentences. The second PLM is BART, specifically ‘bart-large-mnli’, in which a probability threshold was set to 0.5. It took the texts and the list of topics as input and did not require further training or fine-tuning. The performance evaluation showed overall relatively higher AUC scores across most topics, which indicates that this PLM has a better class discrimination ability than BERT for the majority of topics but not better than the traditional method (TF-IDF + Lin-SVM (Classifier Chains)). However, the quality of the resulting F1 scores is less consistent. The traditional method ranks first in terms of performance, followed by BART and BERT. However, despite its superior metrics, its supervised nature makes this method less scalable. These results set a valuable baseline to compare the efficacity of more complex methods and approaches. It is worth mentioning that HuggingFace was resorted to for both PLMs. 
\\[0.25cm]
Before contrasting QUAD-LLM-MLTC to GPT-4o 0-shot, GPT-4o In-Context Learning (ICL) binary evaluation summary, for the largest set (i.e., 1,000) is first investigated (Table 3). This learning includes examples of abstract sentences and their corresponding topics in the prompt. Three cases of few-shot learning were experimented with, i.e., 1-shot, 3-shot, and 5-shot, because in practice it is recommended not to exceed five to six examples. The F1 and AUC scores were highly improved compared to the PLMs results summarized in Table 2, and they were achieved without further training or prior labeling as required by the traditional method. It can be noticed that the topic with the most counts, ‘Sustaining proliferative signaling,’ achieved significantly better performance using supervised learning compared to in-context learning, which indicates that few-shot learning can fall short when the array of topics is relatively nuanced topics, i.e., not easy to identify. This remark is confirmed by the last topic in terms of frequency, where the in-context learning performed significantly better than the supervised learning. This suggests that basic prompt engineering, even with in-context learning, varies significantly based on topic complexity and identifiability, rather than just label balance within the dataset, as is often the case in supervised learning. Additionally, it is noticed that there is no positive correlation between the number of examples provided in the prompt and the approach’s performance. Besides, the AUC scores are high across most topics except ‘Sustaining proliferative signaling’. In-context learning results significantly improve the topic-level performance, in which no labeling is required. This proof of concept demonstrates the potential for a new approach that leverages GPT-4o while enhancing performance across topics and maintaining reliability, even when dealing with varying data volumes.

\begin{table}[H]
\captionsetup{labelformat=empty}
\centering
\renewcommand{\arraystretch}{1.2}
\caption{\textbf{Table 3.} In-Context Learning Results}
\begin{tabular}{c c c c c c c}
\hline
\textbf{In-Context Learning}& \multicolumn{2}{c}{\makecell{\textbf{GPT-4o 1-shot}}} & \multicolumn{2}{c}{\textbf{GPT-4o 3-shot}} & \multicolumn{2}{c}{\textbf{GPT-4o 5-shot}} \\
\hline
 \textbf{Topics/Metrics} & \textbf{F1} & \textbf{AUC} & \textbf{F1} & \textbf{AUC} & \textbf{F1} & \textbf{AUC} \\
\hline
\multicolumn{1}{l}{\textbf{Sustaining proliferative signaling}} & 53.39\%&	67.71\%	&63.47\%	&73.40\%	&63.72\%	&73.60\% \\
\hline
\multicolumn{1}{l}{\textbf{Resisting cell death}} & 86.42\%&	89.58\%	&86.71\%	&89.99\%	&85.48\%	&89.71\% \\
\hline
\multicolumn{1}{l}{\textbf{Activating invasion and metastasis}} & 92.43\%	&94.53\%	&90.81\%	&93.38\%	& 91.62\%	&94.12\% \\
\hline
\multicolumn{1}{l}{\textbf{Genomic instability and mutation}} & 86.49\%	&90.71\%	&86.79\%	&91.29\%	&87.47\%	&92.02\% \\
\hline
\multicolumn{1}{l}{\textbf{Tumor promoting inflammation}} & 65.64\%	& 74.85\%	& 64.86\%	& 74.58\%& 	68.89\%	& 77.24\% \\
\hline
\multicolumn{1}{l}{\textbf{Inducing angiogenesis}} & 94.65\%	&97.12\%	&94.65\%	&97.12\%	&94.17\%	&96.68\% \\
\hline
\multicolumn{1}{l}{\textbf{Evading growth suppressors}} & 58.12\%	&78.97\%	&51.09\%	&70.75\%	&50.27\%	&70.74\% \\
\hline
\multicolumn{1}{l}{\textbf{Enabling replicative immortality}} & 73.10\%	& 79.61\%	& 74.83\%& 	81.03\%	& 58.46\%	& 71.37\% \\
\hline
\multicolumn{1}{l}{\textbf{Avoiding immune destruction}} & 76.13\%	& 88.20\%	& 77.71\%	& 90.02\%	& 80.52\%	& 90.91\% \\
\hline
\multicolumn{1}{l}{\textbf{Cellular energetics}} & 82.09\%	& 98.73\%	& 81.48\%	& 98.67\%	& 79.71\%	& 98.51\% \\
\hline
\end{tabular}
\end{table}

Tables 4, 5, 6, and 7 compare GPT-4 0-shot and QUAD-LLM-MLTC for each dataset, as well as the overall results showing the average and standard deviation of the metrics.  Across all three sets, QUAD-LLM-MLTC consistently outperforms GPT-4o 0-shot for the majority of topics.

\newpage

\begin{table}[H]
\captionsetup{labelformat=empty}
\centering
\renewcommand{\arraystretch}{1.2}
\caption{\textbf{Table 4.} GPT-4o 0-shot vs QUAD-LLM-MLTC Results (Set size = 300)}
\begin{tabular}{c c c c c}
\hline
\textbf{Approach (Set size = 300)}& \multicolumn{2}{c}{\makecell{\textbf{GPT-4o 0-shot}}} &  \multicolumn{2}{c}{\textbf{QUAD-LLM-MLTC}} \\
\hline
 \textbf{Topics/Metrics} & \textbf{F1} & \textbf{AUC} & \textbf{F1} & \textbf{AUC}  \\
\hline
\multicolumn{1}{l}{\textbf{Sustaining proliferative signaling}} &	55.81\%	& 69.32\%	& \textbf{74.84\%}	& 81.33\%\\
\hline
\multicolumn{1}{l}{\textbf{Resisting cell death}} &	92.31\%	& 94.81\%	& \textbf{92.94\%}	& 95.40\%\\
\hline
\multicolumn{1}{l}{\textbf{Activating invasion and metastasis}} &	\textbf{90.57\%}	& 93.63\%	& 85.44\%	& 89.72\%\\
\hline
\multicolumn{1}{l}{\textbf{Genomic instability and mutation}} &	81.43\%	& 85.94\%	& \textbf{85.31\%}	& 88.79\%\\
\hline
\multicolumn{1}{l}{\textbf{Tumor promoting inflammation}} &	63.92\%	& 73.63\%	& \textbf{65.38\%}	& 75.09\%\\
\hline
\multicolumn{1}{l}{\textbf{Inducing angiogenesis}} &	\textbf{94.12\%}	& 98.01\%	& \textbf{94.12\%}	& 98.01\%\\
\hline
\multicolumn{1}{l}{\textbf{Evading growth suppressors}} &	53.33\%	& 75.63\%	& \textbf{58.18\%}	& 76.55\%\\
\hline
\multicolumn{1}{l}{\textbf{Enabling replicative immortality}} &	\textbf{69.77\%}	& 79.45\%	& 58.54\%	& 73.27\%\\
\hline
\multicolumn{1}{l}{\textbf{Avoiding immune destruction}} &	63.16\%	& 82.24\%	& \textbf{69.09\%}	& 84.71\%\\
\hline
\multicolumn{1}{l}{\textbf{Cellular energetics}} &	66.67\%	& 87.89\%	& \textbf{75.86\%}	& 88.59\% \\
\hline
\end{tabular}
\end{table}

\begin{table}[H]
\captionsetup{labelformat=empty}
\centering
\renewcommand{\arraystretch}{1.2}
\caption{\textbf{Table 5.} GPT-4o 0-shot vs QUAD-LLM-MLTC Results (Set size = 500)}
\begin{tabular}{c c c c c}
\hline
\textbf{Approach (Set size = 500)}& \multicolumn{2}{c}{\makecell{\textbf{GPT-4o 0-shot}}} &  \multicolumn{2}{c}{\textbf{QUAD-LLM-MLTC}} \\
\hline
 \textbf{Topics/Metrics} & \textbf{F1} & \textbf{AUC} & \textbf{F1} & \textbf{AUC}  \\
\hline
\multicolumn{1}{l}{\textbf{Sustaining proliferative signaling}} &	58.96\% &	70.62\% &	\textbf{71.66\%} &	78.77\%\\
\hline
\multicolumn{1}{l}{\textbf{Resisting cell death}} &	\textbf{88.06\%} &	91.84\% &	86.67\% &	91.06\%\\
\hline
\multicolumn{1}{l}{\textbf{Activating invasion and metastasis}} &	\textbf{95.88\%} &	98.21\% &	\textbf{95.88\%} &	98.21\%\\
\hline
\multicolumn{1}{l}{\textbf{Genomic instability and mutation}} &	83.65\% &	88.37\% &	\textbf{83.96\%} &	89.04\%\\
\hline
\multicolumn{1}{l}{\textbf{Tumor promoting inflammation}} &	\textbf{64.86\%} &	74.95\% &	63.95\% &	74.41\%\\
\hline
\multicolumn{1}{l}{\textbf{Inducing angiogenesis}} &	\textbf{96.30\%} &	98.78\% &	\textbf{96.30\% }&	98.78\%\\
\hline
\multicolumn{1}{l}{\textbf{Evading growth suppressors}} &	\textbf{51.28\%} &	75.89\% &	49.48\% &	71.44\%\\
\hline
\multicolumn{1}{l}{\textbf{Enabling replicative immortality}} &	71.64\% &	79.05\% &	\textbf{74.29\%} &	81.38\%\\
\hline
\multicolumn{1}{l}{\textbf{Avoiding immune destruction}} &	\textbf{76.54\%} &	92.67\% &	72.73\% &	88.49\%\\
\hline
\multicolumn{1}{l}{\textbf{Cellular energetics}} &	81.36\% &	98.84\%	 &\textbf{81.48\%} &	94.99\%\\
\hline
\end{tabular}
\end{table}

\begin{table}[H]
\captionsetup{labelformat=empty}
\centering
\renewcommand{\arraystretch}{1.2}
\caption{\textbf{Table 6.} GPT-4o 0-shot vs QUAD-LLM-MLTC Results (Set size = 1,000)}
\begin{tabular}{c c c c c}
\hline
\textbf{Approach (Set size = 1,000)}& \multicolumn{2}{c}{\makecell{\textbf{GPT-4o 0-shot}}} &  \multicolumn{2}{c}{\textbf{QUAD-LLM-MLTC}} \\
\hline
 \textbf{Topics/Metrics} & \textbf{F1} & \textbf{AUC} & \textbf{F1} & \textbf{AUC}  \\
\hline
\multicolumn{1}{l}{\textbf{Sustaining proliferative signaling}} &	54.91\%	&68.76\%	&\textbf{74.07\%}&	80.42\%\\
\hline
\multicolumn{1}{l}{\textbf{Resisting cell death}} &	85.04\%	&87.82\%	&\textbf{88.05\%}&	91.28\%\\
\hline
\multicolumn{1}{l}{\textbf{Activating invasion and metastasis}} &	84.27\%&	87.51\%	&\textbf{91.84\%}	&94.75\% \\
\hline
\multicolumn{1}{l}{\textbf{Genomic instability and mutation}} &	85.09\%&	89.80\%	&\textbf{87.83\%}&	92.16\%\\
\hline
\multicolumn{1}{l}{\textbf{Tumor promoting inflammation}} &	\textbf{65.10\%}&	74.29\%	&64.66\%	&74.69\%\\
\hline
\multicolumn{1}{l}{\textbf{Inducing angiogenesis}} &	\textbf{94.56\%}&	96.08\%	&92.56\%&	95.45\%\\
\hline
\multicolumn{1}{l}{\textbf{Evading growth suppressors}} &	57.39\%&	78.13\%	&\textbf{67.27\%}&	82.95\%\\
\hline
\multicolumn{1}{l}{\textbf{Enabling replicative immortality}} &	70.00\%	&77.42\%	&\textbf{75.34\%}	&80.79\%\\
\hline
\multicolumn{1}{l}{\textbf{Avoiding immune destruction}} &	\textbf{78.48\%}&	90.20\%	&76.62\%	&88.25\%\\
\hline
\multicolumn{1}{l}{\textbf{Cellular energetics}} &	81.82\%	&97.87\%&	\textbf{89.08\%}	&97.60\%\\
\hline
\end{tabular}
\end{table}

'Sustaining proliferative signaling', 'Genomic instability and mutation', and 'Cellular energetics' are the topics in which the proposed approach has consistently achieved better performance compared to GPT-4o 0-shot across all three sets. Additionally, based on the conducted experiments, it can be deduced that the performance of the approach is not correlated with the studied set size. This can be justified by the minimal training required in the QUAD-LLM-MLTC. Despite the inconsistency of each topic's performance, the overall multi-label classification performance achieved by the proposed approach is higher than the GPT-4o 0-shot. Additionally, each topic's performance is not influenced by the dataset's size.

\begin{table}[H]
\captionsetup{labelformat=empty}
\centering
\renewcommand{\arraystretch}{1.2}
\caption{\textbf{Table 7.} GPT-4o 0-Shot vs QUAD-LLM-MLTC Results (Overall)}
\begin{tabular}{c c c c c}
\hline
\textbf{Approach Overall}& \multicolumn{2}{c}{\makecell{\textbf{GPT-4o 0-shot}}} &  \multicolumn{2}{c}{\textbf{QUAD-LLM-MLTC}} \\
\hline
 \textbf{Topics/Metrics} & \makecell{\textbf{F1} \\  \textbf{(Mean ± SD)}} & \makecell{\textbf{AUC} \\  \textbf{(Mean ± SD)}} & \makecell{\textbf{F1} \\  \textbf{(Mean ± SD)}} & \makecell{\textbf{AUC} \\  \textbf{(Mean ± SD)}}  \\
\hline
\multicolumn{1}{l}{\textbf{Sustaining proliferative signaling}} &	56.56\% ± 0.021	& 69.57\% ± \textbf{0.010}	& \textbf{73.52\% }± \textbf{0.017}	& \textbf{80.17\%} ± 0.013 \\
\hline
\multicolumn{1}{l}{\textbf{Resisting cell death}} &	88.47\% ± 0.036	& 91.49\% ± 0.035	& \textbf{89.22\%} ± \textbf{0.033}	& \textbf{92.58\%} ± \textbf{0.024}\\
\hline
\multicolumn{1}{l}{\textbf{Activating invasion and metastasis}} &	90.24\% ± 0.058	&93.12\% ± 0.054	&\textbf{91.05\% }± \textbf{0.053}	&\textbf{94.23\% }± \textbf{0.043}\\
\hline
\multicolumn{1}{l}{\textbf{Genomic instability and mutation}} &	83.39\% ± \textbf{0.018}	&88.04\% ± 0.020	& \textbf{85.70\%} ± 0.020&	\textbf{90.00\%} ± \textbf{0.019}\\
\hline
\multicolumn{1}{l}{\textbf{Tumor promoting inflammation}} &	64.63\% ± \textbf{0.006}	&74.29\% ± 0.007	&\textbf{64.66\%} ± 0.007&	\textbf{74.73\%} ± \textbf{0.003}\\
\hline
\multicolumn{1}{l}{\textbf{Inducing angiogenesis}} &	\textbf{94.99\%} ± \textbf{0.012}	& \textbf{97.62\% }± \textbf{0.014}	& 94.33\% ± 0.019& 	97.41\% ± 0.017\\
\hline
\multicolumn{1}{l}{\textbf{Evading growth suppressors}} &	54.00\% ± \textbf{0.031}	& 76.55\% ± \textbf{0.014}	& \textbf{58.31\%} ± 0.089	& \textbf{76.98\%} ± 0.058\\
\hline
\multicolumn{1}{l}{\textbf{Enabling replicative immortality}} &	\textbf{70.47\%} ± \textbf{0.010}	& \textbf{78.64\%} ± \textbf{0.011}	& 69.39\% ± 0.094	& 78.48\% ± 0.045\\
\hline
\multicolumn{1}{l}{\textbf{Avoiding immune destruction}} &	72.73\% ± 0.083	& \textbf{88.37\%} ± 0.054	& \textbf{72.81\%} ± \textbf{0.038}	& 87.15\% ± \textbf{0.021}\\
\hline
\multicolumn{1}{l}{\textbf{Cellular energetics}} &	76.61\% ± 0.086	& \textbf{94.87\%} ± 0.061	& \textbf{82.14\%} ± \textbf{0.066}	& 93.73\% ± \textbf{0.046}\\
\hline
\end{tabular}
\end{table}

The approach takes GPT-4-To 0-shot as the foundation model. It improves the topics' performance by designing specific prompts that include additional context provided by other LLMs (i.e., BERT, PEGASUS) and getting the topics' assignments probabilities using another LLM (i.e., BART). Subsequently, these outputs are fed to a meta-classifier. Therefore, the approach's performance surpasses GPT-4-Turbo 0-shot across the majority of topics in terms of both the metrics' respective means and standard deviations, with two exceptions for which the difference is minimal. The metrics' statistics were calculated based on the performance evaluation of the three sampled sets (i.e., 300, 500, and 1,000 texts). These experiments also validate the reliability of the proposed approach and its consistent improvement across all topics. Moreover, QUAD-LLM-MLTC significantly enhances the previous least performing topic: 'Sustaining proliferative signaling'. The results in Table 7 and Figures 9 and 10 reflect the means and standard deviations (error bars) of the F1 and AUC scores per topic based on the three evaluation sets.

\begin{figure}[H]
    \centering
    \includegraphics[scale=1]{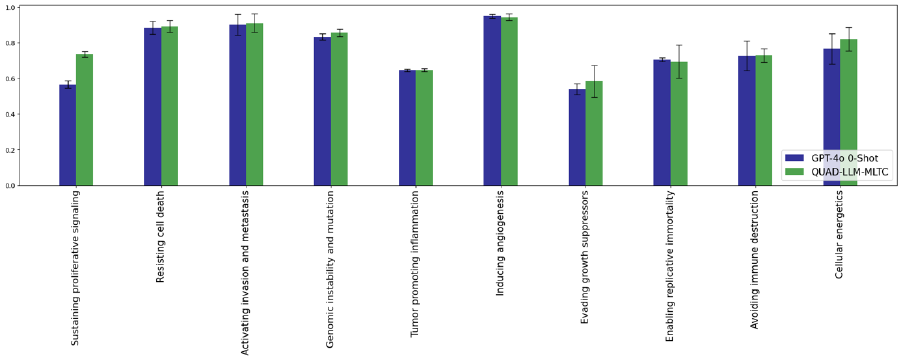}
    \caption*{\textbf{Figure 9.} Model Performance Comparison: F1 scores Mean and Standard Deviation}
    \label{fig:mltc-approaches}
\end{figure}

\newpage 

\begin{figure}[H]
    \centering
    \includegraphics[scale=1]{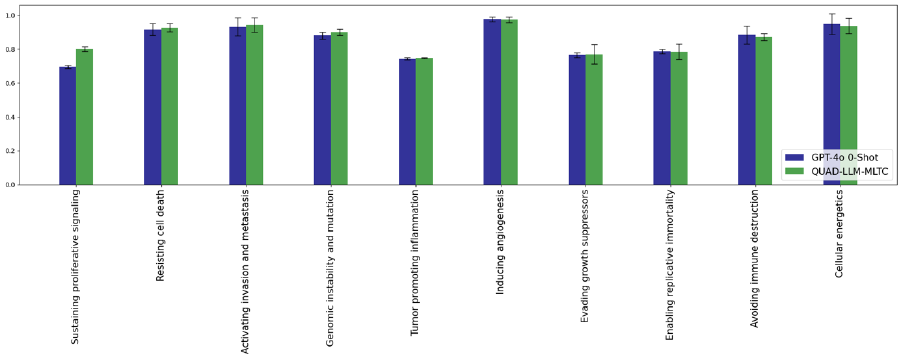}
    \caption*{\textbf{Figure 10.} Model Performance Comparison: AUC scores Mean and Standard Deviation}
    \label{fig:mltc-approaches}
\end{figure}

The comparison bar plots in Figures 9 and 10 are also ordered by the number of texts in which each topic is mentioned based on the largest dataset (i.e., 1,000). The count number decreases from left to right. The majority of the ten topics noticed an improvement with the exception of two in the mean of the F1 score and the exception of four in the mean of the AUC score. Additionally, the variation was reduced across half of the topics. The increased variability in the QUAD-LLM-MLTC approach for some of the topics can be attributed to a combination of reasons. These include the inherent complexity of these cancer hallmarks, potential diversity in the dataset for these topics, and possible trade-offs between improved performance and consistency. The more significant variability in 'Enabling replicative immortality' suggests this topic may be particularly affected by these factors, potentially due to its specific characteristics or representation in the dataset. Therefore, the plots reveal the overall robustness of the proposed approach and suggest that some topics, judged more identifiable by the approach, might lead to less variability. 

\subsubsection{Example-based and Label-based Evaluation}

Because each text in MLTC can be associated simultaneously with multiple topics, it is crucial to adopt multiple evaluation perspectives to understand and compare the approaches' performances thoroughly. After assessing each topic as a separate binary classification problem, it can be noted that this evaluation alone can lead to overlooking the approach's ability to deal with the complexity of assigning multiple labels to each instance. For this reason, the methodologies are also evaluated from example-based and label-based angles. These evaluations are based on the F1 score and its variations because they balance the corresponding precision and recall, which provides more comprehensive performance measures. As a result, the combination of all three types of evaluations enables a holistic view of the proposed approach's performance.

Table 8 shows the evolution of the methods that can be used for MLTC from traditional machine learning to combining multiple advanced LLMs through ensemble learning and highlights a significant positive jump in performance without requiring prior annotation for training. On the one hand, BERT and BART reveal limited performance, in which BERT shows particularly low scores. On the other hand, TF-IDF + Lin-SVM (Classifier Chains) indicates a significant and more consistent improvement across all the considered metrics but still necessitates labeling. GPT-4o displays robust performance across its 0-shot, 1-shot, 3-shot, and 5-shot scenarios while unsupervised. These results demonstrate this LLM's contextual solid understanding. It is worth noting that the traditional machine learning approach and BERT metrics results from a five-fold iterative stratified cross-validation on the largest set (i.e., 1,000). The results reported for BART, the three in-context learning and the 0-shot scenarios are based on the largest dataset as well. Remarkably, QUAD-LLM-MLTC outperforms all the others; it achieves the highest scores across all metrics. Therefore, this approach can better handle this specific MLTC task by leveraging the stacking capability to provide better and more robust performance in an automated manner without manual annotation, except for the meta-classifier training. Additionally, this table demonstrates the promises that more sophisticated models such as LLMs hold for MLTC.

\newpage

\begin{table}[H]
\captionsetup{labelformat=empty}
\centering
\renewcommand{\arraystretch}{1.2}
\caption{\textbf{Table 8.} Example-based and Label-based Results}
\begin{tabular}{c c c c c c}
\hline
& &	\textbf{Example-based} & \multicolumn{3}{c}{\textbf{Label-based}} \\
\hline
\textbf{Category} & \textbf{Approach/Metrics} &	\textbf{F1}	& \textbf{Micro F1}	& \textbf{Macro F1}	& \textbf{Weighted F1} \\
\hline
\textbf{Traditional ML} & \makecell{\textbf{TF-IDF + Lin-SVM} \\ \textbf{(Classifier Chains)}}	&76.63\%	&80.84\%	&77.57\%	&80.43\%\\
\hline
\multirow{2}{*}{\makecell{\textbf{PLM}}} &  \textbf{BERT}	&14.69\%	&17.38\%&	13.33\%&	16.92\%\\
\cmidrule(lr){2-6}
& \textbf{BART}	&28.87\%	&32.29\%	&33.68\%	&36.48\%\\
\hline
\multirow{2}{*}{\makecell{\textbf{ICL}}} & \textbf{GPT-4o 1-Shot}&	74.77\%&	74.99\%&	74.56\%&	73.87\%\\
\cmidrule(lr){2-6}
& \textbf{GPT-4o 3-Shot}&	76.29\%	&78.06\%&	77.24\%&	77.24\%\\
\cmidrule(lr){2-6}
&\textbf{ GPT-4o 5-Shot}&	76.42\%&	77.67\%&	76.03\%&	76.77\%\\
\hline
\textbf{LLM} & \textbf{GPT-4o 0-Shot}&	75.25\%&	75.56\%&	75.67\%&	74.46\%\\
\hline
\multicolumn{2}{l}{\textbf{QUAD-LLM-MLTC (Size = 300)}}	&76.22\%	&79.57\%&	75.97\%	&78.97\%\\
\hline
\multicolumn{2}{l}{\textbf{QUAD-LLM-MLTC (Size = 500)}}	&77.23\%&	79.46\%	&77.64\%&	78.87\%\\
\hline
\multicolumn{2}{l}{\textbf{QUAD-LLM-MLTC (Size = 1,000)}}	&\textbf{81.05\%}	&\textbf{81.45\%}	&\textbf{80.73\%}	& \textbf{80.94\%} \\
\hline
\multicolumn{2}{l}{\textbf{QUAD-LLM-MLTC (Overall)}}	&78.17\%	& 80.16\%	&78.11\%&	79.59\%\\
\hline
\end{tabular}
\end{table}

\subsection{Statistical Validation}

To confirm the robustness of the proposed approach, QUAD-LLM-MLTC, each MLTC approach previously discussed was replicated five times. Table 9 summarizes the descriptive statistics of this analysis. QUAD-LLM-MLTC achieved the highest mean F1 score at 79.28\%, outperforming all the other approaches. Additionally, its strong consistency is shown with a relatively low Standard Deviation (Std Dev) of 0.93\%. The Traditional Machine Learning approach, here TF-IDF + Lin-SVM (Classifier Chains), shows the second-best performance at 76.58\%. Besides, it demonstrates remarkable consistency with the lowest standard deviation (i.e., 0.34\%) among all approaches. Moreover, the performance across different shot settings (i.e., 0, 1, 3, and 5) is relatively consistent, ranging from 71.50\% to 74.57\%. The 3-shot setting performed best among GPT-4 variants at 74.57\%. Interestingly, increasing shots beyond three didn't improve performance. However, 0-shot and in-context learning show relatively higher standard deviations. As expected, based on the previous results, BERT performed surprisingly poorly at 15.04\%. Furthermore, BART showed consistent but low performance at 28.87\% with zero standard deviation, where the same outputs were generated regardless of the replication. In terms of the Confidence Interval (CI), QUAD-LLM-MLTC demonstrates a tight confidence interval [78.13\% - 80.43\%], indicating reliable performance. As a result, QUAD-LLM-MLTC not only improves upon the traditional approach by about 2.7\% but also overcomes the training challenge and labeling requirement, making it an efficient approach for real-world healthcare cases.

\begin{table}[H]
\captionsetup{labelformat=empty}
\centering
\renewcommand{\arraystretch}{1.2}
\caption{\textbf{Table 9.} F1 score Replications Descriptive Statistics}
\begin{tabular}{c c c c c c c}
\hline
\textbf{Approach}	&\textbf{Mean}	&\textbf{Std Dev}&	\textbf{Min}	&\textbf{Max}	
&\textbf{95\% CI Lower}	& \textbf{95\% CI Upper} \\
\hline
\textbf{Traditional ML}&	76.58\%&	\textbf{0.34\%}	&75.98\%&	76.83\%&	76.15\%	&77.00\%\\
\hline
\textbf{BERT}	&15.04\%	&1.42\%&	13.03\%&	16.95\%	&13.27\%&	16.81\%\\
\hline
\textbf{BART}	&28.87\%&	0.00\%	&28.87\%	&28.87\%&	-	&-\\
\hline
\textbf{GPT-4o 1-shot}&	71.50\%&	1.89\%&	70.37\%&	74.77\%&	69.15\%	&73.85\%\\
\hline
\textbf{GPT-4o 3-shot}	&74.57\%&	0.97\%	&73.99\%&	76.29\%&	73.36\%&	75.77\%\\
\hline
\textbf{GPT-4o 5-shot}&	74.15\%&	1.30\%	&73.31\%&	76.42\%&	72.53\%	&75.76\%\\
\hline
\textbf{GPT-4o 0-shot}&	72.80\%&	1.44\%&	71.46\%&	75.25\%&	71.02\%	&74.59\%\\
\hline
\textbf{QUAD-LLM-MLTC}&	\textbf{79.28\%}&	0.93\%	&\textbf{78.32\%}&	\textbf{80.75\%}&	\textbf{78.13\%}&	\textbf{80.43\%} \\
\hline
\end{tabular}
\end{table}

\newpage

Comparing the different approaches based on the corresponding box plots, as shown in Figure 11, confirms the superiority and consistency of the proposed approach, QUAD-LLM-MLTC, in terms of F1 score.                  

\begin{figure}[H]
    \centering
    \includegraphics[scale=1]{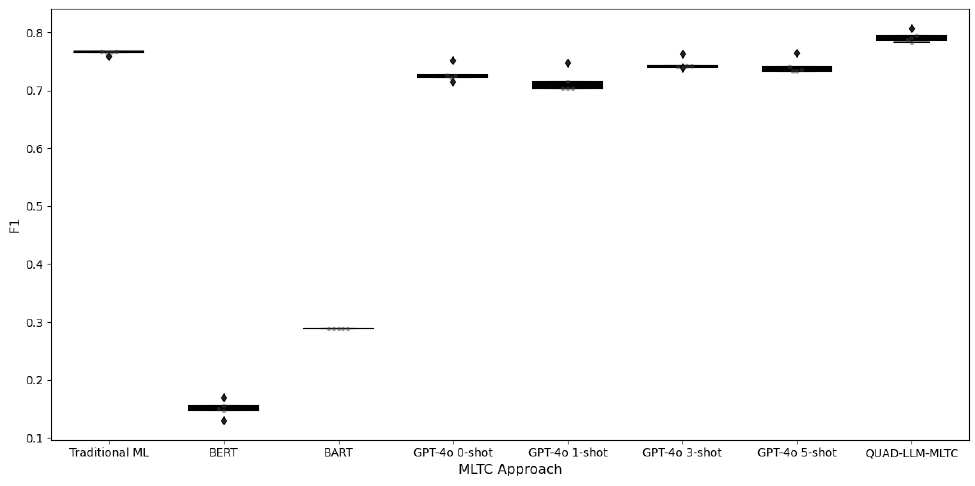}
    \caption*{\textbf{Figure 11.} MLTC Approaches Comparison: Box plots}
    \label{fig:mltc-approaches}
\end{figure}

t-tests demonstrated statistical significance between QUAD-LLM-MLTC and all other MLTC approaches, with p-values lower than 0.05, indicating that the observed differences in F1 scores were not due to randomness, as summarized in Table 10. BERT showed the largest negative mean difference (-0.64), while TF-IDF + Lin-SVM (Classifier Chains) showed the smallest (-0.03).

\begin{table}[H]
\captionsetup{labelformat=empty}
\centering
\renewcommand{\arraystretch}{1.2}
\caption{\textbf{Table 10.} F1 score Replications Descriptive Statistics}
\begin{tabular}{c c c c c c c}
\hline
\textbf{Approach}&	\textbf{Mean Difference}&	\textbf{t-statistic}&	\textbf{p-value}	&\textbf{Significance} \\
\hline
\textbf{BERT}&	-0.64&	-84.63&	0.00&	True \\
\hline
\textbf{BART}&	-0.50&	-121.78&	0.00&	True\\
\hline
\textbf{GPT-4o 1-shot}&	-0.08&	-8.25&	0.00&	True\\
\hline
\textbf{GPT-4o 0-shot}&	-0.06&	-8.46&	0.00&	True\\
\hline
\textbf{GPT-4o 5-shot}&	-0.05&	-7.20&	0.00&	True\\
\hline
\textbf{GPT-4o 3-shot}&	-0.05&	-7.87&	0.00&	True\\
\hline
\textbf{Traditional ML}&	-0.03&	-6.14&	0.00&	True\\
\hline
\end{tabular}
\end{table}

One-way ANOVA was performed, and the resulting metrics, summarized in Table 11, are interesting. The F-statistic of 2190.18 is extremely high, indicating very large differences between the MLTC approaches being compared. This suggests the variance between different approaches is much larger than the variance within the same approach replications. Additionally, these differences are statistically significant. The eta-squared value of 1 reflects a strong effect size, indicating that 100\% of the variance in F1 scores is explained by the different MLTC approaches used. This high value is confirmed by the clear separations between approaches in the box plot, suggesting that the differences in F1 scores are entirely due to the inherent capabilities of each MLTC approach rather than random variations or experimental noise.

\newpage

\begin{table}[H]
\captionsetup{labelformat=empty}
\centering
\renewcommand{\arraystretch}{1.2}
\caption{\textbf{Table 11.} ANOVA Results}
\begin{tabular}{c c c c c c c}
\hline
\textbf{Metric}	& \textbf{Value} \\
\hline
\textbf{F-statistic}	& 2190.18\\
\hline
\textbf{p-value}& 	0.00\\
\hline
\(\bm{\eta}^{2} \) & 	1.00\\
\hline
\end{tabular}
\end{table}

\subsection{Ablation Study}
Tables 12, 13, 14, and 15 summarize the impact of the systematic modifications and combinations that led to proposing QUAD-LLM-MLTC in terms of both the example-based and label-based evaluations. The first three tables reflect the results for each of the considered datasets, while the last one reflects the overall by taking the averages. An increase in performance, as also shown in Table 8, is demonstrated as the dataset size increases, reaching 80\%. The findings suggest that adding contextual elements to the prompt does not consistently improve performance; rather, an integrated and hybrid approach proves more effective.

\begin{table}[H]
\captionsetup{labelformat=empty}
\centering
\renewcommand{\arraystretch}{1.2}
\caption{\textbf{Table 12.} Performance Comparison of Approaches (F1 scores) (Set size = 300)}
\begin{tabular}{c c c c c c c}
\hline
\textbf{Approach (Size = 300)}	& \textbf{F1}	& \textbf{Micro F1}	& \textbf{Macro F1}	& \textbf{Weighted F1} \\
\hline
\textbf{Classification 1}	& 73.39\%	& 76.17\%	& 73.11\%	& 75.11\% \\
\hline
\textbf{Classification 2}	& 70.40\%	& 72.73\%	& 70.99\%	& 71.88\% \\
\hline
\textbf{Classification 3}	& 70.65\%	& 73.14\%	& 71.18\%	& 72.04\% \\
\hline
\textbf{Hard Voting}	& 71.29\%& 	73.92\%	& 72.71\%	& 72.71\% \\
\hline
\textbf{QUAD-LLM-MLTC}	& \textbf{76.22\%}	& \textbf{79.57\%}	& \textbf{75.97\%}	& \textbf{78.97\%} \\
\hline
\end{tabular}
\end{table}

\begin{table}[H]
\captionsetup{labelformat=empty}
\centering
\renewcommand{\arraystretch}{1.2}
\caption{\textbf{Table 13.} Performance Comparison of Approaches (F1 scores) (Set size = 500)}
\begin{tabular}{c c c c c c c}
\hline
\textbf{Approach (Size = 500)}	& \textbf{F1}	& \textbf{Micro F1}	& \textbf{Macro F1}	& \textbf{Weighted F1}\\
\hline
\textbf{Classification 1}& 	75.54\%	& 77.75\%	& 76.85\%	& 76.76\%\\
\hline
\textbf{Classification 2}& 	71.98\%	& 73.87\%	& 72.92\%	& 72.98\%\\
\hline
\textbf{Classification 3}& 	71.35\%	& 73.48\%	& 72.39\%& 	72.32\%\\
\hline
\textbf{Hard Voting	}&  72.76\%& 	75.00\%& 	74.41\%	& 73.77\%\\
\hline
\textbf{QUAD-LLM-MLTC}&  \textbf{77.23\%}	& \textbf{79.46\%}	& \textbf{77.64\%}	& \textbf{78.87\%} \\
\hline
\end{tabular}
\end{table}

\begin{table}[H]
\captionsetup{labelformat=empty}
\centering
\renewcommand{\arraystretch}{1.2}
\caption{\textbf{Table 14.} Performance Comparison of Approaches (F1 scores) (Set size = 1,000)}
\begin{tabular}{c c c c c c c}
\hline
\textbf{Approach (Size = 1,000)}	& \textbf{F1}	& \textbf{Micro F1}	& \textbf{Macro F1}	& \textbf{Weighted F1}\\
\hline
\textbf{Classification 1}	&75.25\%&	75.56\%&	75.67\%&	74.46\%\\
\hline
\textbf{Classification 2}	&75.65\%	&76.23\%	&75.34\%	&75.29\%\\
\hline
\textbf{Classification 3}	&74.00\%	&74.99\%	&74.56\%	&73.87\%\\
\hline
\textbf{Hard Voting}	&72.62\%	&74.25\%&	73.40\%	&72.66\%\\
\hline
\textbf{QUAD-LLM-MLTC}	&\textbf{81.05\%}	&\textbf{81.45\%}	&\textbf{80.73\%}&	\textbf{80.94\%} \\
\hline
\end{tabular}
\end{table}

Interestingly, the inconsistent reduction in performance among Classification 1, 2, and 3 reflects a significant loss of topic accuracy across different classifications and a corresponding gain in accuracy for other topics because the customized aggregation of these classifications with additional adjustments led to a more robust approach. In parallel, these three classifications were also combined through hard voting, and the results are reported in the different ablation study tables. As a result, no significant improvement can be discussed. This research study reveals the importance of carefully choosing the relevant components to include in the crafted prompts approach and validating it through multiple experiments using different sets of textual data (i.e., abstracts' sentences). Furthermore, it highlights the promise of leveraging various LLMs to provide sufficient context in the prompt engineering process. These findings propose that tailoring the classification strategies to the specific characteristics of each label (i.e., topic) and combining them through ensemble learning can significantly improve accuracy. 

\begin{table}[H]
\captionsetup{labelformat=empty}
\centering
\renewcommand{\arraystretch}{1.2}
\caption{\textbf{Table 15.} Performance Comparison of Approaches (F1 scores) (Overall)}
\begin{tabular}{c c c c c}
\hline
\textbf{Approach (Overall)} &	\makecell{\textbf{F1} \\ \textbf{(Mean ± SD)}}	& \makecell{\textbf{Micro F1} \\ \textbf{(Mean ± SD)}}& \makecell{\textbf{Macro F1} \\ \textbf{(Mean ± SD)}}	&\makecell{\textbf{Weighted F1} \\ \textbf{(Mean ± SD)}} \\
\hline
\textbf{Classification 1}	&74.73\% ± 0.012&	76.49\% ± 0.011	&75.21\% ± 0.019	&75.44\% ± 0.012\\
\hline
\textbf{Classification 2}	&72.68\% ± 0.027	&74.28\% ± 0.018	&73.08\% ± 0.022	&73.39\% ± 0.017\\
\hline
\textbf{Classification 3}	&72.00\% ± 0.018&	73.87\% ± 0.010	&72.71\% ± 0.017	&72.74\% ± 0.010\\
\hline
\textbf{Hard Voting}&	72.22\% ± \textbf{0.008}&	74.39\% ± \textbf{0.006}	&73.51\% ± \textbf{0.009}&	72.68\% ± \textbf{0.000}\\
\hline
\textbf{QUAD-LLM-MLTC}	&\textbf{78.17\%} ± 0.025	&\textbf{80.16\%} ± 0.011	&\textbf{78.11\%} ± 0.024	&\textbf{79.59\%} ± 0.012\\
\hline
\end{tabular}
\end{table}

\section{Conclusion and Future Directions}
This research proposes a novel approach, QUAD-LLM-MLTC, which combines four LLMS—GPT-4o, BERT, PEGASUS, and BART—followed by Stacking to conduct MLTC for healthcare textual data. This research demonstrates the efficacy of integrating multiple methods to provide more context to the prompt and handle the complexities and challenges faced when conducting an MLTC. This approach surpasses traditional and single-model methodologies regarding the F1 score while ensuring consistency. It is characterized by a sequential pipeline in which BERT is used to extract the text’s key tokens, PEGASUS to augment the textual data and provide two text variations, GPT-4o to carry out the classification, and BART to provide topics’ assignment probabilities. The ensemble method stacking provides an additional layer to the proposed approach, which achieves better performance than any base classifier across all the topics.
\\[0.25cm]
Despite the promising results, this research has limitations that must be acknowledged. Conducting any text classification using LLMs requires constant and recurrent human feedback for evaluation. This ensures alignment and the guarantee of the classification’s good performance. Although LLMs’ application enables scalability due to the lack of required training and extensive annotation, the constant need for human intervention can constitute an obstacle, particularly because the nuances of topics and texts might evolve over time. Furthermore, given the sequential nature of the pipeline, the performance of QUAD-LLM-MLTC is contingent upon the quality of the key tokens extracted and the texts augmented. Therefore, any errors that occur at the first stages could propagate through the pipeline, which would impact the classification’s accuracy. 
\\[0.25cm]
Looking ahead, several research lines can be explored to further the QUAD-LLM-MLTC approach framework. First, the expansion of the experiments to a wider variety of topics and healthcare datasets would allow a more targeted and generalized improvement. Additionally, although computationally expensive and resource demanding, exploring fine-tuning the LLM responsible for classification on the specific dataset subject to study can lead to the precise weight adjustments tailored for the classification task and potentially enhance the approach’s ability to recognize the different labels more accurately.

\section*{References}
\begin{enumerate}[label={[\arabic*]}, leftmargin=*]
    \item 	Nam, J., Kim, J., Loza Mencía, E., Gurevych, I., \& Fürnkranz, J. (2014). Large-scale multi-label text classification—revisiting neural networks. In Machine Learning and Knowledge Discovery in Databases: European Conference, ECML PKDD 2014, Nancy, France, September 15-19, 2014. Proceedings, Part II 14 (pp. 437-452). Springer Berlin Heidelberg.
    \item 	Zhao, W. X., Zhou, K., Li, J., Tang, T., Wang, X., Hou, Y., Min, Y., Zhang, B., Zhang, J., Dong, Z., Du, Y., Yang, C., Chen, Y., Chen, Z., Jiang, J., Ren, R., Li, Y., Tang, X., Liu, Z., Liu, P., Nie, J.-Y., \& Wen, J.-R. (2024). A survey of large language models. arXiv. https://arxiv.org/abs/2303.18223.
    \item 	Brown, T. B., Mann, B., Ryder, N., Subbiah, M., Kaplan, J., Dhariwal, P., Neelakantan, A., Shyam, P., Sastry, G., Askell, A., Agarwal, S., Herbert-Voss, A., Krueger, G., Henighan, T., Child, R., Ramesh, A., Ziegler, D. M., Wu, J., Winter, C., Hesse, C., Chen, M., Sigler, E., Litwin, M., Gray, S., Chess, B., Clark, J., Berner, C., McCandlish, S., Radford, A., Sutskever, I., \& Amodei, D. (2020). Language models are few-shot learners. arXiv. https://arxiv.org/abs/2005.14165.
    \item 	Loukas, L., Stogiannidis, I., Diamantopoulos, O., Malakasiotis, P., \& Vassos, S. (2023, November). Making LLMs worth every penny: Resource-limited text classification in banking. In Proceedings of the Fourth ACM International Conference on AI in Finance (pp. 392-400).
    \item 	Zhang, Y., Wang, M., Ren, C., Li, Q., Tiwari, P., Wang, B., \& Qin, J. (2024). Pushing The Limit of LLM Capacity for Text Classification. arXiv preprint arXiv:2402.07470.
    \item 	Devlin, J., Chang, M. W., Lee, K., \& Toutanova, K. (2018). Bert: Pre-training of deep bidirectional transformers for language understanding. arXiv preprint arXiv:1810.04805.
    \item 	Yang, Z., Dai, Z., Yang, Y., Carbonell, J., Salakhutdinov, R. R., \& Le, Q. V. (2019). Xlnet: Generalized autoregressive pretraining for language understanding. Advances in neural information processing systems, 32.
    \item 	Zaheer, M., Guruganesh, G., Dubey, K. A., Ainslie, J., Alberti, C., Ontañón, S., Pham, P., Ravula, A., Wang, Q., Yang, L., \& Ahmed, A. (2020). Big Bird: Transformers for longer sequences. Advances in Neural Information Processing Systems, 33, 17283–17297.
    \item 	Beltagy, I., Peters, M. E., \& Cohan, A. (2020). Longformer: The long-document transformer. arXiv preprint arXiv:2004.05150.
    \item 	Yin, W., Hay, J., \& Roth, D. (2019). Benchmarking zero-shot text classification: Datasets, evaluation and entailment approach. arXiv preprint arXiv:1909.00161.
    \item 	Chalkidis, I., Fergadiotis, M., Malakasiotis, P., Aletras, N., \& Androutsopoulos, I. (2020). LEGAL-BERT: The muppets straight out of law school. arXiv preprint arXiv:2010.02559.
    \item 	Jiang, Z., Xu, F. F., Araki, J., \& Neubig, G. (2020). How can we know what language models know?. Transactions of the Association for Computational Linguistics, 8, 423-438.
    \item 	Zhang, J., Zhao, Y., Saleh, M., \& Liu, P. (2020, November). PEGASUS: Pre-training with extracted gap-sentences for abstractive summarization. In International conference on machine learning (pp. 11328-11339). PMLR.
    \item 	Lewis, M., Liu, Y., Goyal, N., Ghazvininejad, M., Mohamed, A., Levy, O., Stoyanov, V., \& Zettlemoyer, L. (2020). BART: Denoising sequence-to-sequence pre-training for natural language generation, translation, and comprehension. In Proceedings of the 58th Annual Meeting of the Association for Computational Linguistics (pp. 7871-7880). Association for Computational Linguistics.
    \item 	Wang, Y., Wang, L., Rastegar-Mojarad, M., Moon, S., Shen, F., Afzal, N., Liu, S., Zeng, Y., Mehrabi, S., Sohn, S., \& Liu, H. (2018). Clinical information extraction applications: A literature review. Journal of Biomedical Informatics, 77, 34-49.
    \item 	Sakai, H., Mikaeili, M., Lam, S. S., \& Bosire, J. (2023). Text Classification for Patient Experience Improvement: A Neural Network Approach. In IIE Annual Conference. Proceedings (pp. 1-6). Institute of Industrial and Systems Engineers (IISE).
    \item 	Mujtaba, G., Shuib, L., Idris, N., Hoo, W. L., Raj, R. G., Khowaja, K., Shaikh, K., \& Nweke, H. F. (2019). Clinical text classification research trends: systematic literature review and open issues. Expert systems with applications, 116, 494-520.
    \item 	Tsoumakas, G., \& Katakis, I. (2007). Multi-label classification: An overview. International Journal of Data Warehousing and Mining (IJDWM), 3(3), 1-13.
    \item 	Zhang, M. L., \& Zhou, Z. H. (2013). A review on multi-label learning algorithms. IEEE transactions on knowledge and data engineering, 26(8), 1819-1837.
    \item 	Ghali, M. K., Farrag, A., Sakai, H., Baz, H. E., Jin, Y., \& Lam, S. (2024). GAMedX: Generative AI-based medical entity data extractor using large language models. arXiv preprint arXiv:2405.20585.
    \item 	Vithanage, D., Deng, C., Wang, L., Yin, M., Alkhalaf, M., Zhang, Z., Zhu, Y., Soewargo, A. C., \& Yu, P. (2024). Evaluating machine learning approaches for multi-label classification of unstructured electronic health records with a generative large language model. medRxiv, 2024-06.
    \item 	Li, R., Wang, X., \& Yu, H. (2024, May). LlamaCare: An Instruction Fine-Tuned Large Language Model for Clinical NLP. In Proceedings of the 2024 Joint International Conference on Computational Linguistics, Language Resources and Evaluation (LREC-COLING 2024) (pp. 10632-10641).
    \item 	Bumgardner, V. C., Mullen, A., Armstrong, S. E., Hickey, C., Marek, V., \& Talbert, J. (2024). Local Large Language Models for Complex Structured Tasks. AMIA Summits on Translational Science Proceedings, 2024, 105.
    \item 	Gema, A. P., Minervini, P., Daines, L., Hope, T., \& Alex, B. (2023). Parameter-efficient fine-tuning of LLaMA for the clinical domain. arXiv preprint arXiv:2307.03042.
    \item 	Guo, Z. (2024). Automated Text Mining of Experimental Methodologies from Biomedical Literature. arXiv preprint arXiv:2404.13779.
    \item 	Bansal, P., Das, S., Rai, V., \& Kumari, S. (2023). Multi-label Classification of Covid-19 Vaccine Tweet.
    \item 	Zhang, Y., Li, X., Liu, Y., Li, A., Yang, X., \& Tang, X. (2023). A multilabel text classifier of cancer literature at the publication level: methods study of medical text classification. JMIR Medical Informatics, 11(1), e44892.
    \item 	Uslu, E. E., Sezer, E., \& Guven, Z. A. (2024). NLP-Powered Insights: A Comparative Analysis for Multi-Labeling Classification with MIMIC-CXR Dataset. IEEE Access.
    \item 	Yogarajan, V., Montiel, J., Smith, T., \& Pfahringer, B. (2021, June). Transformers for multi-label classification of medical text: an empirical comparison. In International Conference on Artificial Intelligence in Medicine (pp. 114-123). Cham: Springer International Publishing.
    \item 	Guevara, M., Chen, S., Thomas, S., Chaunzwa, T. L., Franco, I., Kann, B. H., Moningi, S., Qian, J. M., Goldstein, M., Harper, S., Aerts, H. J. W. L., Catalano, P. J., Savova, G. K., Mak, R. H., \& Bitterman, D. S. (2024). Large language models to identify social determinants of health in electronic health records. npj Digital Medicine, 7, 6.
    \item 	Bețianu, M., Mălan, A., Aldinucci, M., Birke, R., \& Chen, L. (2024, April). DALLMi: Domain Adaption for LLM-Based Multi-label Classifier. In Pacific-Asia Conference on Knowledge Discovery and Data Mining (pp. 277-289). Singapore: Springer Nature Singapore.
    \item 	Ray, S., Mehta, P., Zhang, H., Chaman, A., Wang, J., Ho, C. J., Chiou, M., \& Suleman, T. (2023). Segmented Harmonic Loss: Handling Class-Imbalanced Multi-Label Clinical Data for Medical Coding with Large Language Models. arXiv preprint arXiv:2310.04595.
    \item 	Chen, Q., Du, J., Allot, A., \& Lu, Z. (2022). LitMC-BERT: transformer-based multi-label classification of biomedical literature with an application on COVID-19 literature curation. IEEE/ACM transactions on computational biology and bioinformatics, 19(5), 2584-2595.
    \item 	Nguyen, B., \& Ji, S. (2021). Fine-Tuning Pretrained Language Models With Label Attention for Biomedical Text Classification. arXiv preprint arXiv:2108.11809.
    \item 	Ge, X., Williams, R. D., Stankovic, J. A., \& Alemzadeh, H. (2023). DKEC: Domain knowledge enhanced multi-label classification for electronic health records. arXiv preprint arXiv:2310.07059.
    \item 	Kirkpatrick, J., Pascanu, R., Rabinowitz, N., Veness, J., Desjardins, G., Rusu, A. A., Milan, K., Quan, J., Ramalho, T., Grabska-Barwinska, A., Hassabis, D., Clopath, C., Kumaran, D., \& Hadsell, R. (2017). Overcoming catastrophic forgetting in neural networks. Proceedings of the National Academy of Sciences, 114(13), 3521-3526.
    \item 	Strubell, E., Ganesh, A., \& McCallum, A. (2020, April). Energy and policy considerations for modern deep learning research. In Proceedings of the AAAI conference on artificial intelligence (Vol. 34, No. 09, pp. 13693-13696).
    \item 	Sushil, M., Zack, T., Mandair, D., Zheng, Z., Wali, A., Yu, Y. N., Quan, Y., \& Butte, A. J. (2024). A comparative study of zero-shot inference with large language models and supervised modeling in breast cancer pathology classification. Research Square.
    \item 	Sakai, H., Lam, S. S., Mikaeili, M., Bosire, J., \& Jovin, F. (2024). Large Language Models for Patient Comments Multi-Label Classification. arXiv preprint arXiv:2410.23528.
    \item 	Zhu, C. Q., Stureborg, R., \& Dhingra, B. (2024). Hierarchical Multi-Label Classification of Online Vaccine Concerns. arXiv preprint arXiv:2402.01783.
    \item 	Sarkar, S., Feng, D., \& Santu, S. K. K. (2023, December). Zero-shot multi-label topic inference with sentence encoders and LLMs. In Proceedings of the 2023 Conference on Empirical Methods in Natural Language Processing (pp. 16218-16233).
    \item 	Raja, H., Munawar, A., Delsoz, M., Elahi, M., Madadi, Y., Hassan, A., Abu Serhan, H., Inam, O., Hermandez, L., Tran, S., Munir, W., Abd-Alrazaq, A., Chen, H., \& Yousefi, S. (2024). Using Large Language Models to Automate Category and Trend Analysis of Scientific Articles: An Application in Ophthalmology. arXiv preprint arXiv:2308.16688.
    \item 	Lehman, E., Hernandez, E., Mahajan, D., Wulff, J., Smith, M. J., Ziegler, Z., Nadler, D., Szolovits, P., Johnson, A., \& Alsentzer, E. (2023). Do We Still Need Clinical Language Models? Proceedings of Machine Learning Research, 209, 578-597.
    \item 	Amin, S., Neumann, G., Dunfield, K., Vechkaeva, A., Chapman, K. A., \& Wixted, M. K. (2019, September). MLT-DFKI at CLEF eHealth 2019: Multi-label Classification of ICD-10 Codes with BERT. In CLEF (Working Notes) (pp. 1-15).
    \item 	Yogarajan, V., Pfahringer, B., Smith, T., \& Montiel, J. (2022, September). Concatenating BioMed-Transformers to Tackle Long Medical Documents and to Improve the Prediction of Tail-End Labels. In International Conference on Artificial Neural Networks (pp. 209-221). Cham: Springer Nature Switzerland.
    \item 	Kementchedjhieva, Y., \& Chalkidis, I. (2023). An exploration of encoder-decoder approaches to multi-label classification for legal and biomedical text. arXiv preprint arXiv:2305.05627.
    \item 	Chaichulee, S., Promchai, C., Kaewkomon, T., Kongkamol, C., Ingviya, T., \& Sangsupawanich, P. (2022). Multi-label classification of symptom terms from free-text bilingual adverse drug reaction reports using natural language processing. PLoS One, 17(8), e0270595.
    \item Baker, S., Silins, I., Guo, Y., Ali, I., Högberg, J., Stenius, U., \& Korhonen, A. (2016). Automatic semantic classification of scientific literature according to the hallmarks of cancer. Bioinformatics, 32(3), 432-440.
    \item 	OpenAI. (2024a, May 13). Hello GPT-4o. OpenAI. https://openai.com/index/hello-gpt-4o/
    \item 	OpenAI. (2024b). Enterprise privacy. OpenAI. https://openai.com/enterprise-privacy/
    \item 	Breiman, L. (2001). Random forests. Machine learning, 45, 5-32.
    \item 	Friedman, J. H. (2001). Greedy function approximation: a gradient boosting machine. Annals of statistics, 1189-1232.
    \item 	Freund, Y., \& Schapire, R. E. (1997). A decision-theoretic generalization of on-line learning and an application to boosting. Journal of computer and system sciences, 55(1), 119-139.
    \item 	Dietterich, T. G. (2000, June). Ensemble methods in machine learning. In International workshop on multiple classifier systems (pp. 1-15). Berlin, Heidelberg: Springer Berlin Heidelberg.
    \item 	Wolpert, D. H. (1992). Stacked generalization. Neural networks, 5(2), 241-259.
    \item 	Sorower, M. S. (2010). A literature survey on algorithms for multi-label learning. Oregon State University, Corvallis, 18(1), 25.

\end{enumerate}

\end{document}